\documentclass[11pt]{article}
\usepackage{graphicx}
\usepackage[margin=1.0in]{geometry}
\usepackage{standalone}
\usepackage{amsfonts}
\usepackage{amsmath}
\usepackage{amssymb}
\usepackage{tikz-cd}
\usepackage{hyperref}
\usepackage{blindtext}
\usepackage{enumitem}
\usepackage{braket}
\usepackage{mathtools}
\usepackage{environ}
\usepackage{fancyhdr}
\usepackage{stmaryrd}
\usepackage{tikz-cd}
\usepackage{xcolor}
\usepackage{pgfplots}
\usepackage{multirow}
\usepackage{tikz}
\usepackage{colortbl}
\usepackage{comment}
\usepackage{verbatim}
\usepackage{caption}
\usepackage{booktabs}
\usepackage{float}
\usepackage{subcaption}
\usepackage{bbm}
\usepackage[blocks]{authblk}
\restylefloat{table}

\usepackage[mathlines]{lineno}

\allowdisplaybreaks
\newtheorem{theorem}{Theorem}[section]
\newtheorem{proposition}[theorem]{Proposition}
\newtheorem{lemma}[theorem]{Lemma}
\newtheorem{corollary}[theorem]{Corollary}
\newtheorem{definition}[theorem]{Definition}
\newtheorem{example}[theorem]{Example}
\newtheorem{remark}[theorem]{Remark}
\newtheorem{condition}[theorem]{Condition}

\let\olddefinition\definition
\let\oldexample\example
\let\oldremark\remark
\let\oldcondition\condition

\renewcommand{\definition}{\olddefinition\normalfont}
\renewcommand{\example}{\oldexample\normalfont}
\renewcommand{\remark}{\oldremark\normalfont}
\renewcommand{\condition}{\oldcondition\normalfont}
\newenvironment{proof}{\noindent{\bf Proof:}}{$\hfill \Box$ \vspace{10pt}}  

\newtheorem{assumption}{Assumption}

\newcommand\numberthis{\addtocounter{equation}{1}\tag{\theequation}}

\DeclareMathOperator\sort{sort}
\DeclareMathOperator\diam{diam}
\DeclareMathOperator\proj{proj}

\usepackage[backend=bibtex, 
            style=numeric,giveninits=true, maxbibnames=99,hyperref=true,sortcites=true]{biblatex}
\title{Approximating invariant functions with the sorting trick is theoretically justified}
\author{Wee Chaimanowong\thanks{The Chinese University of Hong Kong.}\quad{}\quad{} Ying Zhu\thanks{University of California San Diego.}}

\date{}

\begin{document}
\maketitle

\begin{abstract}
Many machine learning models leverage group invariance which is enjoyed with a wide-range of applications. For exploiting an invariance structure, one common approach is known as \emph{frame averaging}. One popular example of frame averaging is the \emph{group averaging}, where the entire group is used to symmetrize a function. Another example is the \emph{canonicalization}, where a frame at each point consists of a single group element which transforms the point to its orbit representative, for example, sorting. Compared to group averaging, canonicalization is more efficient computationally. However, it results in non-differentiability or discontinuity of the canonicalized function. As a result, the theoretical performance of canonicalization has not been given much attention. In this work, we establish an approximation theory for canonicalization. Specifically, we bound the point-wise and $L^2(\mathbb{P})$ approximation errors as well as the eigenvalue decay rates associated with a canonicalization trick applied to reproducing kernels. We discuss two key insights from our theoretical analyses and why they point to an interesting future research direction on how one can choose a design to fully leverage canonicalization in practice.
\end{abstract}


\bigskip


\section{Introduction}
Given a group $G$ acting on a set $\mathcal{X}$ and a function $f: \mathcal{X}\rightarrow \mathcal{Y}$, $f$ is called $G$-\emph{invariant} if $f(g\cdot x) = f(x)$ for all $g\in G$ and $x\in X$.
Many machine learning models leverage group invariance which is enjoyed with a wide-range of applications such as set predictions,
point-cloud classification and segmentation,
graph neural network for molecular classification, and social networks analysis.
A number of options are available in the literature for exploiting an invariance structure. 
One common approach is known as \emph{frame averaging} \cite{dym2024equivariant, puny2021frame}. The idea of frame averaging is to specify a suitable set-valued function $\mathcal{X}\rightarrow 2^G$, known as a frame, so that a function can be symmetrized by averaging over all the transformed inputs from the action of elements in a frame. 
One example of frame averaging is the \emph{group averaging}, where the entire group is used to symmetrize a function. Another example is the \emph{canonicalization}, where a frame at each point consists of a single group element which transforms the point to its orbit representative, for example, sorting. The motivation for canonicalization is, averaging over the entire group can be computationally expensive, especially when the group cardinality is large. An active research topic is to compute, learn, and/or approximate an efficient frame which ensures desirable properties for the frame-averaged function. 

\bigskip

\begin{figure}[ht!]
    \centering
    \includegraphics[width=0.4\linewidth]{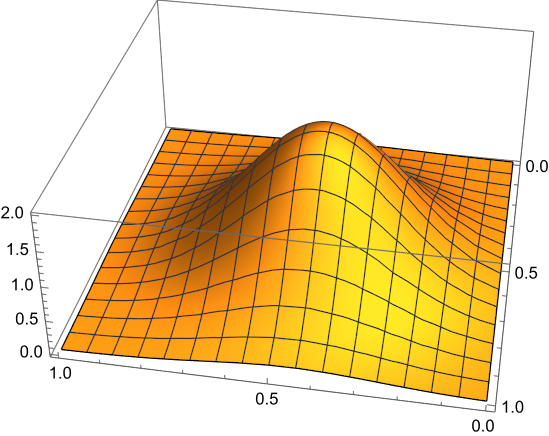}
    \includegraphics[width=0.4\linewidth]{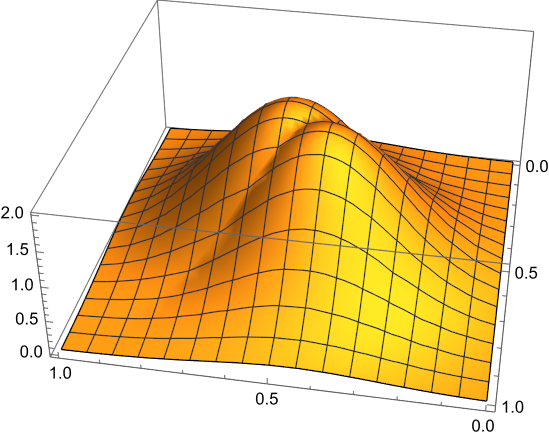}
    \caption{A smooth Gaussian kernel $\mathcal{K}(w,z) = \frac{1}{2\pi}\exp\left(-\frac{1}{2}\|w-z\|_2^2\right)$ over $\mathcal{X} = [0,1]^2$ centered at $z = (0.0,0.2)$ (left), and the canonicalized version $\mathcal{K}^{\sort}(w,z) = \frac{1}{2\pi}\exp\left(-\frac{1}{2}\|\sort w-\sort z\|_2^2\right)$ (right). The kernel $\mathcal{K}^{\sort}$ is permutation invariant and continuous, but not differentiable.}
    \label{fig:sortedkernel}
\end{figure}

In this work, we consider approximating a real-valued $G$-invariant function from an RKHS on $\mathcal{X}$ via a canonicalized kernel interpolation. When $G$ is the permutation group $S_d$ acting on $\mathcal{X} = [0,1]^d$ by the coordinate permutation 
\begin{equation*}
    x := (x^1,\cdots,x^d) \in \mathcal{X} \mapsto \sigma x = (x^{\sigma(1)}, \cdots, x^{\sigma(d)}) \in \mathcal{X}
\end{equation*}
where the superscripts denote the coordinate indices, a canonicalized kernel in our context takes the form $\mathcal{K}(\sort ., \sort .)$. 
Little attention has been given to the theoretical performance of canonicalization in approximation. One possible reason is the poor analytical properties of the canonicalization map, leading to non-differentiability or discontinuity of the canonicalized function \cite{dym2024equivariant, zhang2019fspool}. To illustrate, consider a Gaussian kernel $\mathcal{K}(w,z) = \frac{1}{2\pi}\exp\left(-\frac{1}{2}\|w-z\|_2^2\right)$ over $[0,1]^2$, which is smooth. But $\mathcal{K}^{\sort}(w,z) := \mathcal{K}(\sort w, \sort z)$, invariant under the action of the permutation group $G = S_2$ by coordinate permutation, is non-differentiable at $w = z$; see Figure \ref{fig:sortedkernel}. 

We show that despite poor analytical properties may arise from canonicalization, certain theoretical performance guarantees can be derived. These results strengthen the obvious computational benefit a canonicalization approach has over other computationally intensive frame-averaging.
There is a subtle difference between our sorting trick and the common way canonicalization is used in the literature. The latter sorts the evaluation point of a permutation invariant function $f(\cdot)$. The sorting trick we consider applies to a reproducing kernel. When it is used together with data points $\{y_i,x_i\}_{i=1}^{n}$ where $y_i=f(x_i)$ to approximate $f(\cdot)$, \emph{both} the data and evaluation points are sorted, i.e., $\mathcal{K}(\sort x, \sort x_i)$. One key insight from our theoretical justification lies in the \emph{rearrangement principle}: if $w^1\geq\cdots\geq w^d$ and $z^1\geq\cdots\geq z^d$, then $w^1z^1+\cdots+w^dz^d \geq w^1z^{\sigma(1)}+\cdots+w^dz^{\sigma(d)}$ for every permutation $\sigma \in S_d$. Intuitively, viewing $w$ as an evaluation point and $z$ as a data point, the sorted evaluation point is always closer to the sorted data point compared to their unsorted counterparts. Therefore, $\mathcal{K}(\sort w, \sort z)$ improves approximation by evaluating closer to the interpolated points.

\bigskip

We use tools from approximation theory \cite{wendland2004scattered} to show the improvement in the upper bounds for the point-wise and $L^2(\mathbb{P})$ approximation errors relative to the standard kernel without sorting. 
Our insight from the rearrangement principle is that: by mapping sample points to the fundamental domain $\mathcal{X}^{\sort}$ of $\mathcal{X}$, sorting reduces the fill distance, a crucial concept in the theory of interpolation in mathematics. We also show that there exists a design sequence $X = \{x_i\}_{i=1}^{n} \subset \mathcal{X}$ such that the upper bound for the $L^2(\mathbb{P})$ approximation error associated with the sorted kernel is reduced by a factor of $\sqrt{(d!)^{\nu/d}}$, where $\nu$ denotes the smoothness degree of the kernel. This result prompts an interesting future research question on how one can choose a design sequence in $\mathcal{X}$ to fully take advantage of the sorting technique. 

\bigskip
We find that our point-wise approximation error bound is weaker near the boundary of the fundamental domain, because of the more restrictive \emph{interior cone conditions}. This result is a manifestation of the loss of the kernel's differentiability. However, we show that as the number of sample points $n$ becomes large, the effect of the weaker approximation near the boundary on the $L^2(\mathbb{P})$-norm diminishes, as the measure of the area considered `near' the boundary approaches zero. Lastly,  we bound the sorted kernel's eigenvalue decay rate. It is known that the eigenvalue decay rate of a $C^\nu$-smoothness kernel is $\lesssim j^{\nu/d}$. Our result shows that the sorting trick reduces the bound by a factor of $(d!)^{\nu/d-1}$ as long as $\nu/d > 1$.

\bigskip
Our analysis in this work focuses on a permutation group $S_d$ acting on $\mathcal{X} = [0,1]^d$ to illustrate the key point. However, the extension to an arbitrary finite group $G$ acting on a compact domain $\mathcal{X}\hookrightarrow \mathbb{R}^d$ equipped with a $G$-invariant probability measure with a bounded density is straightforward. For example, our results for the approximation errors continue to hold after adjusting the statements to accommodate $|\mathcal{X}/G| = \sum_{g\in G}|\mathcal{X}^g|/|G|$ by Burnside's Lemma, replacing the factor of $1/d!$. Here, $\mathcal{X}^g$ denotes the set of points in $\mathcal{X}$ invariant under the action of $g\in G$, and we assume that the fundamental domain $\mathcal{X}/G$ is embedded in $\mathcal{X}$. Therefore, $\mathcal{X}/G\hookrightarrow \mathcal{X}$ plays the role of the domain $\mathcal{X}^{\sort}\hookrightarrow \mathcal{X}$ of sorted elements in the case of $G = S_d$. 

\bigskip 

\textbf{Related Works.} Many studies have conducted experiments on the performance of frame averaging and canonicalization \cite{dym2024equivariant,puny2021frame,du2022se,kaba2023equivariance,duval2023faenet,zhang2018learning,ma2024canonicalization, steinert2025integration}. The theoretical guarantees of performance improvement in invariant learning have also been studied in various settings. Earlier work such as \cite{shawe1989building} is concerned with the reduction in the trainable degree of freedom for the invariant neural network. \cite{ginsbourger2012argumentwise} formalizes the notion of invariant Reproducing Kernel Hilbert Spaces (RKHS) and Gaussian process regressions given an arbitrary invariant kernel. More recent work shows reduction in the generalization errors for empirical risk minimization over a given invariant function class \cite{petrache2023approximation,sannai2021improved,sokolic2017generalization}. A number of papers specialize in an RKHS symmetrized with a group-averaged kernel \cite{bietti2024sample, mei2021learning, elesedy2021provably}. Meanwhile, canonicalization has gathered some attention \cite{dym2024equivariant, zhang2019fspool, aslan2022group}: In particular, \cite{dym2024equivariant} points out the potential undesirable non-differentiability or discontinuity properties of the canonicalized functions; \cite{aslan2022group} discusses an efficient implementation of the canonicalization map itself, for an arbitrary group symmetry. A few papers have discussed kernel canonicalization where both evaluation point and data point are mapped to the fundamental domain. Notably, a paper relevant to ours is \cite{steinert2025integration}, which introduces a canonicalized kernel in the context of Gaussian process regressions and demonstrates the performance improvement \emph{empirically} on datasets from quantum chemistry and ocean science. In contrast, our work introduces the canonicalized kernel, formalizes the corresponding RKHS, and leverages approximation theory to \emph{theoretically} justify the improved performances of the canonicalized methods people have observed empirically despite the poor analytical properties. In recent years, approximation theoretic tools \cite{wendland2004scattered} have been used to obtain error bounds of kernel regressions in settings outside invariance \cite{tuo2020improved, wang2020prediction}, and from a broader perspective, our work belongs to this stream of literature.

\bigskip

\textbf{Notation.} We use superscripts for coordinate indices and subscripts are reserved for sample indices, i.e., $x = (x^1,\cdots,x^d) \in \mathbb{R}^d$, and $x^a_i$ denotes the $a$th coordinate of the $i$th sample vector $x_i \in \mathbb{R}^d$; $\mathcal{X}$ denotes a compact domain in $\mathbb{R}^d$, and unless specified otherwise, we will focus on $\mathcal{X} := [0,1]^d$. Given a set $X := \{x_i\}_{i=1}^n \subset \mathcal{X}$, for any $\mathcal{K} : \mathcal{X}\times\mathcal{X}\rightarrow \mathbb{R}$, $\mathcal{K}(X,X)$ is the matrix with the $(i,j)$ entry $\mathcal{K}(x_i,x_j)$.  For any $\sigma \in S_d$, we write $\sigma X$ for $\{\sigma x_i\}_{i=1}^n$, and $\sort X$ for $\{\sort x_i\}_{i=1}^n$. The $L^2(\mathcal{X}, \mathbb{P})$-norm $\|f-g\|^2_{L^2(\mathcal{X}, \mathbb{P})} := \int_{\mathcal{X}}|f(x)-g(x)|^2d\mathbb{P}(x)$ and the Euclidean norm 
$ \|x_1-x_2\|_2^2 := \sum_{a=1}^d (x^a_1 - x^a_2)^2
$. Consider a given $\mathcal{X}\subset \mathbb{R}^d$. We denote by $|\mathcal{X}|$ the Euclidean $d$-volume of $\mathcal{X}$. Let $B(x,\varepsilon) := \{x' \in \mathcal{X}\ |\ \|x'-x\|_2\leq \varepsilon\}$ be a Euclidean $d$-ball of radius $\varepsilon > 0$, and let $\omega_d := |B(0,1)|$ be the volume of a unit $d$-ball. We denote by $C^\nu(\mathcal{X})$ the space of $C^\nu$-smooth functions. We will always assume $\nu > 0$ when writing $C^\nu$, and specifically write $C^0(.)$ and $C^\infty(.)$ for the space of continuous functions and infinitely smooth functions, respectively. We use the multivariate notation: Given $\alpha = (\alpha^1,\cdots,\alpha^d) \in \mathbb{Z}^d_{\geq 0}$, $|\alpha| := \sum_{a=1}^d\alpha^a$, $\alpha! := \prod_{a=1}^d\alpha^a!$, $(x-x_i)^\alpha := \prod_{a=1}^d(x^a-x_i^a)^{\alpha^a}$, and $\partial^\alpha := \partial_{x^1}^{\alpha^1}\cdots\partial_{x^d}^{\alpha^d}$.



\section{Sorted RKHS}



\begin{definition}
    A symmetric bivariate function $\mathcal{K}:\mathcal{X}\times \mathcal{X}\rightarrow \mathbb{R}$ is called \emph{positive (semi)definite} if for any given set $X = \{x_i \in \mathcal{X}\}_{i=1}^n$ of \emph{distinct} points, the matrix $\mathcal{K}(X,X)$ is positive (semi)definite.
\end{definition}

We consider a reproducing kernel Hilbert space (RKHS) $\mathcal{H}$ that is equipped with the symmetric positive definite kernel function $\mathcal{K} : \mathcal{X}\times \mathcal{X}\rightarrow \mathbb{R} \in C^\nu(\mathcal{X}\times\mathcal{X})$ for some $\nu \geq 0$.

\bigskip

The permutation invariant subspace $\mathcal{H}^{perm}$ of an RKHS $\mathcal{H}$ is generated by the kernel 
\begin{equation}\label{perminvkernel}
\mathcal{K}^{perm} := \frac{1}{(d!)^2}\sum_{\sigma,\sigma'\in S_d}\mathcal{K}(\sigma ., \sigma' .) 
\end{equation}
Let us consider the case where $\mathcal{X} = [0,1]^d$, equipped with an $S_d$-invariant probability measure $\mathbb{P}$ with a bounded density (with respect to the Lebesgue measure), unless specified otherwise. The symmetrization trick in (\ref{perminvkernel}) can be generalized to an averaging kernel over a finite (or compact) group. This type of symmetrized kernel via averaging 
has been widely studied in various contexts such as \cite{bietti2021sample, mroueh2015learning, haasdonk2005invariance}. It is important to note that (\ref{perminvkernel}) involves a summation over $|S_d|^2 = (d!)^2$ terms, which is computationally expensive. 

\bigskip
If the group action is compatible with the kernel in the sense that $\mathcal{K}(x,y) = \mathcal{K}(\sigma x, \sigma y)$ for all $x, y \in \mathcal{X}$ and $\sigma \in S_d$
,\footnote{This condition holds if $\mathcal{H}$ is invariant under the group $G$, i.e., $f \in \mathcal{H} \implies f(\sigma .) \in \mathcal{H}$ and $\braket{f,g}_{\mathcal{H}} = \braket{f(\sigma .),g(\sigma .)}_{\mathcal{H}}$ for all $\sigma \in G$ (see \cite[Theorem 10.6]{wendland2004scattered}).}
then we have
\begin{equation}\label{perminvkernel_2}
\mathcal{K}^{perm} = \frac{1}{d!}\sum_{\sigma \in S_d}\mathcal{K}(\sigma .,.).
\end{equation}
This compatibility assumption appears in \cite{bietti2021sample, haasdonk2005invariance} but is not used in our analysis. Albeit reducing the computation in (\ref{perminvkernel}),(\ref{perminvkernel_2}) still involves a summation over $|S_d| = d!$ terms. 
 

\bigskip

Instead, the canonicalization approach we propose computues the following kernel that \emph{sorts} the input, 
\begin{equation}\label{sortedkernel}
    \mathcal{K}^{\sort} := \mathcal{K}(\sort ., \sort .),
\end{equation}
where, $\sort : \mathcal{X}\rightarrow \mathcal{X}^{\sort}$ sends any $x = (x^1,\cdots,x^d) \in \mathcal{X}\subset \mathbb{R}^d$ to the representative point \begin{equation}\sort x := (x^{\sigma(1)},\cdots,x^{\sigma(d)})
\end{equation}
in the \emph{fundamental domain} $\mathcal{X}^{\sort} \subset \mathbb{R}^d$, for $\sigma \in S_d$ such that \begin{equation}
x^{\sigma(1)} \geq \cdots \geq x^{\sigma(d)}.
\end{equation}
This operation can be done efficiently in $O(d\log d)$ time.\footnote{Note that the actions are fixed in group averaging while dependent on the inputs in sorting. Therefore, group averaging preserves differentiability while sorting does not.}

\bigskip
To analyze the performance of the sorting trick, we construct the RKHS $\mathcal{H}^{\sort}$ associated with $\mathcal{K}^{\sort}$.\footnote{This discussion is based on \cite[Chapter 10]{wendland2004scattered}, which provides a comprehensive review of native spaces and RKHS construction.} Let us consider
\begin{equation*}
    F_{\mathcal{K}}^{\sort} := span\left\{\mathcal{K}^{\sort}(.,x)\ |\ x\in \mathcal{X}^{\sort}\right\},
\end{equation*}
and define the inner-product on $F_{\mathcal{K}}^{\sort}$ by
\begin{equation*}
    \left\langle \sum_{j=1}^n\alpha_j\mathcal{K}^{\sort}(.,x_i), \sum_{j'=1}^{n'}\alpha'_{j'}\mathcal{K}^{\sort}(.,x'_{j'})\right\rangle_F := \sum_{j=1}^n\sum_{j'=1}^{n'}\alpha_j\alpha'_{j'}\mathcal{K}^{\sort}(x_j,x'_{j'}).
\end{equation*}
Given the set of points $\sort X = \{\sort x_i\}_{i=1}^n$, we have that
\begin{equation*}
    \mathcal{K}^{\sort}(X,X) = \mathcal{K}(\sort X, \sort X)
\end{equation*}
is positive semidefinite when $\mathcal{K}$ is positive semidefinite. Then the inner product $\braket{.,.}_F$ is well defined and the kernel $\mathcal{K}^{\sort}$ satisfies the reproducing properties for the inner product space $F^{\sort}_{\mathcal{K}}$. Note that $F^{\sort}_{\mathcal{K}}$ is not complete and hence not a Hilbert space. To proceed, we consider the completion $\mathcal{F}^{\sort}_{\mathcal{K}}$ of $F^{\sort}_{\mathcal{K}}$, and then we identify $\mathcal{F}^{\sort}_{\mathcal{K}}$ with a subspace $R(\mathcal{F}^{\sort}_{\mathcal{K}})$ of continuous functions, $C^0(\mathcal{X})$. We will check that $R(\mathcal{F}^{\sort}_{\mathcal{K}})$ is the RKHS with kernel $\mathcal{K}^{\sort}$, and hence can be uniquely identified with $\mathcal{H}^{\sort}$ by \cite[theorem 10.11]{wendland2004scattered}. We provide the detail in the following.

\bigskip
Let $\mathcal{F}_{\mathcal{K}}^{\sort}$ be the completion of $F_{\mathcal{K}}^{\sort}$; i.e., the space of equivalent classes of Cauchy sequences in $F_{\mathcal{K}}^{\sort}$ with the inner product $\langle f, g\rangle_{\mathcal{F}} := \lim_{n\rightarrow \infty}\langle f_n, g_n\rangle_F$ for any equivalence classes $f$ and $g$ of the Cauchy sequences $\{f_n\}_{n=1}^\infty$ and $\{g_n\}_{n=1}^\infty$, respectively. 
Let us define a linear map
\begin{equation*}
    R : \mathcal{F}^{\sort}_{\mathcal{K}} \rightarrow C^0(\mathcal{X}), \qquad R(f)(x) := \left\langle f, \mathcal{K}^{\sort}(.,x)\right\rangle_{\mathcal{F}}, \ \forall f \in \mathcal{F}^{\sort}_{\mathcal{K}}.
\end{equation*}
Note that $R(\mathcal{K}^{\sort}(.,x)) : y\mapsto \left\langle \mathcal{K}^{\sort}(.,x), \mathcal{K}^{\sort}(.,y)\right\rangle_{\mathcal{F}} = \mathcal{K}^{\sort}(x,y)$, so we can just identify $R\mathcal{K}^{\sort}(.,x)$ with $\mathcal{K}^{\sort}(.,x)$.

\begin{lemma}\label{rmaplemma}
    The linear map $R$ is injective with the image in $C^0(\mathcal{X})\hookrightarrow L^2(\mathcal{X},\mathbb{P})$.
\end{lemma}
\begin{proof}
    The fact that $R(f)$ is continuous follows from the Cauchy–Schwarz inequality: $\left|R(f)(x) - R(f)(y)\right| = \left\langle f, \mathcal{K}^{\sort}(.,x) - \mathcal{K}^{\sort}(.,y)\right\rangle_{\mathcal{F}} \leq \|f\|_{\mathcal{F}}\cdot \|\mathcal{K}^{\sort}(.,x) - \mathcal{K}^{\sort}(.,y)\|_{F} = \|f\|_{\mathcal{F}}\cdot \sqrt{\mathcal{K}^{\sort}(x,x) + \mathcal{K}^{\sort}(y,y) - 2\mathcal{K}^{\sort}(x,y)}$, and the RHS approaches zero as $x\rightarrow y$ by the continuity of $\mathcal{K}^{\sort}$.

\bigskip
    The injectivity can be seen as follows. Suppose that $f$ is an equivalence class of some Cauchy sequence $\{f_n\}_{n=1}^\infty \subset F^{\sort}_{\mathcal{K}}$.  If $R(f) = 0$, we must have $\|f\|^2_{\mathcal{F}} = \lim_{n\rightarrow \infty}\left\langle f, f_n \right\rangle_{\mathcal{F}} = 0$, proving $f=0$. 
\end{proof}

Define the \emph{native} space
\begin{equation*}
    \mathcal{H}^{\sort} := R(\mathcal{F}^{\sort}_{\mathcal{K}}), \qquad \left\langle R(f), R(g)\right\rangle_{\mathcal{H}^{\sort}} := \left\langle f, g\right\rangle_{\mathcal{F}}.
\end{equation*}
One can see that $\mathcal{H}^{\sort}$ is a Hilbert space as it is an isometric image of one, and most importantly, that $\mathcal{K}^{\sort}$ is the reproducing kernel for $\mathcal{H}^{\sort}$ since $\mathcal{K}^{\sort}(.,x)$ is identified with $R\mathcal{K}^{\sort}(.,x) \in \mathcal{H}^{\sort}$, and $\left\langle R(f), \mathcal{K}^{\sort}(.,x)\right\rangle_{\mathcal{H}^{\sort}} = \left\langle f, \mathcal{K}^{\sort}(.,x)\right\rangle_{\mathcal{F}} = R(f)(x)$. In other words, $\mathcal{H}^{\sort}$ is the RKHS associated with the kernel $\mathcal{K}^{\sort}$, and we will write $R(f) \in \mathcal{H}^{\sort}$ as $f$ for simplicity. From \cite[Theorem 10.11]{wendland2004scattered}, we know that $\mathcal{H}^{\sort}$ is the unique Hilbert space with reproducing kernel $\mathcal{K}^{\sort}$.

\bigskip
Our approximation theory is built upon the problem of minimal norm interpolation of the values $y_i = f(x_i), i=1,\cdots,n$, sampled from an unknown $d$-variate function $f : \mathcal{X}\rightarrow\mathbb{R}$ over the sequence $X = \{x_i\}_{i=1}^n$ of design points. Formally, this problem can be formulated as
\begin{equation*}
    \hat{f}_n \in \arg\min_{\check{f} \in \mathcal{H}}\|\check{f}\|_{\mathcal{H}}, \quad \text{subject to $\check{f}(x_i) = y_i, \ i=1,\cdots,n$},
\end{equation*}
where $\hat{f}_n = \sum_{i=1}^n\hat{\pi}_i\mathcal{K}(.,x_i)$ with $\hat{\pi} := \mathcal{K}(X,X)^{-1}y \in \mathbb{R}^{n}$ if $\mathcal{K}(X,X)$ is positive definite.

\paragraph{Sorting-based approximation.}


Suppose that we observe $n$ observations of an arbitrary unknown permutation invariant $d$-variate function $f:\mathcal{X}\rightarrow \mathbb{R}$ in the form $y_i=f(x_i)$ for $i=1,...,n$. 
We consider the interpolation problem via the sorted RKHS:

\begin{equation}\label{eq:sortedinterpolationproblem}
    \hat{f}_n^{\sort} \in \arg\min_{\check{f} \in \mathcal{H}^{\sort}}\|\check{f}\|_{\mathcal{H}^{\sort}}, \quad \text{subject to} \quad \check{f}(x_i) = y_i, \ i=1,\cdots,n,
\end{equation}
where $\hat{f}_n^{\sort} = \sum_{i=1}^n\hat{\pi}_i^{\sort}\mathcal{K}^{\sort}(., x_i) \in \mathcal{H}^{\sort}$ with $\hat{\pi}^{\sort} := \mathcal{K}^{\sort}(X,X)^{-1}y \in \mathbb{R}^n$, if $\mathcal{K}(X,X)$ is positive definite and $X = \{x_i\}_{i=1}^n$ consists of points in distinct orbits of $S_d$. In particular, for a given $f \in \mathcal{H}^{perm}$, we will later show the convergence of $\hat{f}_n^{\sort}$ to $f$ of the $\mathcal{H}^{\sort}$ interpolant under the $L^2(\mathcal{X}, \mathbb{P})$-norm, subsequentially proving the embedding $\mathcal{H}^{perm}\subset \overline{\mathcal{H}}^{\sort}$, the closure of $\mathcal{H}^{\sort}$ in $L^2(\mathcal{X},\mathbb{P})$.

\begin{remark}
    Note that given a regular kernel $\mathcal{K}\in C^\nu(\mathcal{X}\times \mathcal{X})$, the sorted kernel $\mathcal{K}^{\sort}$ is generally \emph{not} differentiable, although we still have $\mathcal{K}^{\sort}\in C^0(\mathcal{X}\times\mathcal{X})$.

    \bigskip
    
    To generalize permutation invariance, one may consider a finite group $G$ acting on a compact domain $\mathcal{X}\subset \mathbb{R}^d$ with an embedding $\mathcal{X}/G\hookrightarrow \mathcal{X}$. Let us define the map $\pi:\mathcal{X}\rightarrow \mathcal{X}/G\hookrightarrow \mathcal{X}$. Then we can generalize the notion of sorted kernel by $\mathcal{K}^\pi := \mathcal{K}(\pi ., \pi .)$. The construction of the corresponding RKHS $\mathcal{H}^\pi$ follows exactly as we have outlined except that the linear map $R$ codomain is now the set of bounded functions $B(\mathcal{X})\hookrightarrow L^2(\mathcal{X},\mathbb{P})$. While the boundedness follows from the fact that $\mathcal{K}$ is continuous on the compact domain $\mathcal{X}\times\mathcal{X}$, neither $\mathcal{K}^\pi$ nor $R(f)$ are guaranteed to be continuous since $\pi$ might not be continuous in general as proven in \cite{dym2024equivariant}. However, the important fact is that $R$ remains injective, and as a result, we can define $\left\langle R(f), R(g)\right\rangle_{\mathcal{H}^\pi} := \left\langle f, g\right\rangle_{\mathcal{F}}$.
\end{remark}


\section{A theory for sorting-based approximation}\label{approximation theory}

In this section, we establish a theory for the sorting-based approximation using the machinery in \cite{wendland2004scattered}. 

\subsection{Approximation error}

Before presenting our results, let us introduce the key definitions and assumptions. 


\begin{assumption}\label{pd before sort}
    The RKHS $\mathcal{H}$ is generated by the kernel $\mathcal{K} \in C^{\nu}(\mathcal{X}\times \mathcal{X})$, which is positive definite.
    
\end{assumption}


\begin{assumption}\label{PD on data}
    The dataset $X = \{x_i\in \mathcal{X}\}_{i=1}^n$ consists of points in distinct orbits of $S_d$.
\end{assumption}

    If we consider the set of points $\sort X = \{\sort x_i\}_{i=1}^n$, then $\mathcal{K}^{\sort}(X,X)$ is positive semidefinite when $\mathcal{K}$ is positive definite. If we also know that $X$ consists of points in distinct orbits of $S_d$, then $\sort X$ consists of distinct points. In this case, $\alpha^{\intercal}\mathcal{K}^{\sort}(X,X)\alpha = \alpha^{\intercal}\mathcal{K}(\sort X,\sort X)\alpha > 0$ for all $\alpha\in \mathbb{R}^n\setminus\{0\}$ when $\mathcal{K}$ is positive definite. This fact allows us to obtain the unique interpolant $\hat{f}_n^{\sort} = \sum_{i=1}^n\hat{\pi}_i^{\sort}\mathcal{K}^{\sort}(., x_i) \in \mathcal{H}^{\sort}$ where $\hat{\pi}^{\sort} := \mathcal{K}^{\sort}(X,X)^{-1}y$.

    \begin{remark}
        Assumption \ref{PD on data} is not restrictive for most practical purposes. For example, if the data points $X$ are drawn i.i.d. from a distribution with a continuous density, then Assumption \ref{PD on data} holds almost surely.
    \end{remark}

\begin{definition}\label{filldistancedefinition}\textbf{(Fill distance \cite[Definition 1.4]{wendland2004scattered})}
    Given a set of points $X = \{x_i\}_{i=1}^n \subset \mathcal{X} \subset \mathbb{R}^d$, the \emph{fill distance} is given by
    \begin{equation*}
        h_{X, \mathcal{X}} := \sup_{x\in \mathcal{X}}\min_{i=1,\cdots,n}\|x-x_i\|_2,
    \end{equation*}
    where $\|.\|_2$ denotes the Euclidean norm in $\mathbb{R}^d$.
\end{definition}

A simplified overview of the error analysis behind our results in this section suggests a two-step process. First, given a ground-truth function $f$ evaluated at a point $x$, we approximate $f(x)$ with the Taylor's polynomial up to degree $\nu$. Second, we seek a local interpolation of the Taylor's polynomial at $x$ such that this interpolation provides a `good' local approximation of the polynomial. Thus, the approximation error is controlled by the fill distance raised to the power $\nu$. The second step is made possible with the so-called \emph{local polynomial reproduction} stated in Lemma \ref{coneconditionalternativelemma}. The main operational requirement of the local polynomial reproduction at $x$ is the \emph{local interior cone condition} which we describe below.

\color{black}

\begin{definition}\label{interiorconedefinition}\textbf{(Interior cone condition \cite[Definition 3.6]{wendland2004scattered})}
    We say that a set $\Omega \subset \mathbb{R}^d$ satisfies a $(\theta, r)$-interior cone condition for $\theta \in (0, \pi/2)$ and $r > 0$, if for every $x \in \Omega$, there exists a unit vector $\xi_x$ such that the cone
    \begin{equation*}
        C(x,\xi_x, \theta, r) := \left\{x + \tau y\ |\ y \in \mathbb{R}^d, \|y\|_2 = 1, y\cdot \xi_x \geq \cos\theta, \tau \in [0,r]\right\} 
    \end{equation*}
    is contained in $\Omega$.
\end{definition}

\begin{assumption}[Local interior cone condition]\label{cone condition}
    For the given point $x\in \mathcal{X}$ and the dataset $X = \{x_i\}_{i=1}^n$, there exists $\Omega_x\subset\mathcal{X}$ containing $x$ such that $\diam \Omega_x := \sup_{x',x''\in \Omega_x}\|x'-x''\|_2 \leq D$, and $\Omega_x$ satisfies the $(\theta, r)$-interior cone condition for some $\theta \in (0,\pi/2)$ such that 
    \begin{equation}r \geq 4\nu^2\left(1 + \frac{1}{\sin \theta}\right)h, \label{cone_condition_general}
    \end{equation} 
    where
    \begin{equation*}
    h := \sup_{x'\in \Omega_x}\min_{i \in I_x}\|x'-x_i\|_2, \quad   I_{x} := \left\{i\ |\ x_i \in \Omega_x, i=1,\cdots,n\right\}. 
    \end{equation*}
\end{assumption}

We now state Theorem \ref{thm:localapproxerror}, which is a version of \cite[Theorem 11.13]{wendland2004scattered}) that provides a foundation for the rest of our results to built upon.

\begin{theorem}\label{thm:localapproxerror}
    Suppose Assumptions \ref{pd before sort}-\ref{cone condition} are satisfied for the given dataset $X = \{x_i\}_{i=1}^n$ and $x \in \mathcal{X}$. Then 
    \begin{equation*}
        \left|f(x) - \hat{f}_n(x)\right| \leq \|f\|_{\mathcal{H}}\cdot \sqrt{8{\nu + 2d \choose 2d}C_{\mathcal{K},\nu, x, D} \cdot D^{\nu}},\,\textrm{for any } f\in \mathcal{H}
    \end{equation*}
    where 
    \begin{equation*}
        C_{\mathcal{K},\nu, x,D} := \max_{\xi_1, \xi_2 \in \mathcal{X}\cap B(x,D)}\max_{|\alpha|+|\beta|=\nu}\frac{1}{\alpha!\beta !}\left|\partial_1^\alpha\partial_2^\beta\mathcal{K}(\xi_1,\xi_2)\right|.
    \end{equation*}
\end{theorem}

The proof of Theorem \ref{thm:localapproxerror} is given in the appendix. Theorem \ref{thm:localapproxerror} is a modified version of \textbf{\cite[Theorem 11.13]{wendland2004scattered}}. In particular, the modification accommodates the \emph{local} version of the interior cone condition, without requiring an entire domain of interest $\Omega$ to satisfy an \emph{interior cone condition} for some uniform $(\theta, r)$. The key is to apply the following local polynomial reproduction result, a version of \cite[Theorem 3.8 and Theorem 3.14]{wendland2004scattered}, which we also modify to accommodate the \emph{local} interior cone condition. The reason for these modifications will become clear as we present one of our main results, Theorem \ref{prop:hpermembeddedinhsortlocal}.

\begin{lemma}\label{coneconditionalternativelemma}\emph{\textbf{(Local polynomial reproduction)}}
    Suppose Assumption \ref{cone condition} is satisfied for the given dataset $X = \{x_i\}_{i=1}^n$ and $x \in \mathcal{X}$. Then there exist $\{\widetilde{u}_j(x)\}_{j=1}^n$ such that 
    \begin{itemize}
        \item $\sum_{j=1}^n\widetilde{u}_j(x)p(x_j) = p(x)$, for all $p \in \pi_{\nu}(\mathbb{R}^d)$,
        \item $\sum_{j=1}^n\left|\widetilde{u}_j(x)\right| \leq 2$,
        \item $\widetilde{u}_j(x) = 0$ provided that $\|x - x_j\|_2 > D$,
    \end{itemize}
    where $\pi_\nu(\mathbb{R}^d)$ denotes the space of $d$-variates polynomials of degree at most $\nu$. 
\end{lemma}

The self-contained proof of Lemma \ref{coneconditionalternativelemma} is given in the appendix.

\paragraph{Sketch of the Proof of Theorem \ref{thm:localapproxerror}:}

From $\hat{f}_n = \mathcal{K}(.,X)\mathcal{K}(X,X)^{-1}y$, we can show that
    \begin{equation*}
        \left|f(x) - \hat{f}_n(x)\right| \leq \|f\|_{\mathcal{H}}\cdot \sqrt{Q_x(u^*(x))},\numberthis
    \end{equation*}
where $Q_x : \mathbb{R}^n\rightarrow \mathbb{R}$ is given by $Q_x(u) := \mathcal{K}(x,x) - 2\mathcal{K}(x,X)u + u^{\intercal}\mathcal{K}(X,X)u$ and $u^*(x) = \mathcal{K}(X,X)^{-1}\mathcal{K}(X,x) \in \mathbb{R}^n$. It turns out that $u^*(x) \in \mathbb{R}^n$ is a global minimum of $Q_x$. Therefore, we can bound $Q_x(u^*(x))$ by $Q_x(\widetilde{u}(x))$ with $\widetilde{u}(x)$ given in Lemma \ref{coneconditionalternativelemma}, and use the reproducing properties of $\widetilde{u}(x)$ to simplify the Taylor expansion of $\mathcal{K}$ centered at $(x,x)$. Namely, all the lower order terms in $(x-x_i)$, $i=1,\cdots,n$, vanishes. We have:
\begin{multline*}
    Q_x(u^*) \leq Q_x(\widetilde{u}) =  -2\sum_{j=1}^n\sum_{|\alpha|=\nu}\widetilde{u}_j(x)R_{0,\alpha}(x,x;x,x_j)(x-x_j)^\alpha\\
        + \sum_{i,j=1}^n\sum_{|\alpha|+|\beta|=\nu}\widetilde{u}_i(x)\widetilde{u}_j(x)R_{\alpha,\beta}(x,x;x_i,x_j)(x-x_i)^\alpha(x-x_j)^\beta,
\end{multline*}
where $R_{\alpha, \beta}(x_1,x_2;x'_1,x'_2) := \frac{1}{\alpha!\beta!}\partial^\alpha_1\partial^\beta_2\mathcal{K}(\xi_1, \xi_2)$ for some $(\xi_1, \xi_2) \in \mathcal{X}\times \mathcal{X}$ on the line connecting any $(x_1, x_2)\in \mathcal{X}\times \mathcal{X}$ and $(x'_1, x'_2)\in \mathcal{X}\times \mathcal{X}$. From the vanishing property, $\widetilde{u}_j(x) = 0$ if $\|x-x_j\|_2 > D$, we have $\left|x-x_i\right|^\alpha \leq \|x-x_i\|_2^{|\alpha|} \leq D^{|\alpha|}$, and we can bound $|\partial^\alpha_1\partial^\beta_2\mathcal{K}(\xi_1, \xi_2)|$ by $C_{K, \nu, x, D}$ over a compact set $\mathcal{X}\cap B(x, D)$. After some simplifications, we obtain that $Q_x(u^*(x)) \leq 8{\nu + 2d \choose 2d}C_{\mathcal{K},\nu, x,D}D^{2\nu}$. \hfill $\square$

\bigskip

The \emph{local} version of the interior cone condition allows us to handle domains with irregular shapes. For domains with simple geometry, we can find an interior cone with a uniformly large $\theta$ at all points in the domain. As an example, it is known that a $d$-dimensional unit ball satisfies a $(\theta=\pi/3, r = 1)$-interior cone condition, for all $d > 0$ (see \cite[Lemma 3.10]{wendland2004scattered}). Beyond this simple example, unfortunately in most cases, even if we start with a nice $\mathcal{X}$ like a cube, it is inevitable that the fundamental domain $\mathcal{X}/G$ has ``narrow corners'' where the interior cone angle would decay rapidly with higher $d$. Instead of letting the ``narrow corners'' constrain the interior cone condition of the entire domain, we partition it into many small subdomains.
The idea is, most subdomains do not intersect the domain's boundary and allow more relaxed interior cone conditions; hence, the point-wise approximation error bound according to Theorem \ref{thm:localapproxerror} is relatively sharp for most points in the domain. Meanwhile, the approximation error bounds are worse for points in the subdomains that intersect the domain's boundary, which may be interpreted as a pathological manifestation of the canonicalized kernel (i.e., the loss of the kernel’s differentiability). However, the fraction of such boundary subdomains tends to zero as the fill distance tends to zero (as the number of sample points becomes large). 
\bigskip

For our work, the relevant domains are $\mathcal{X} = [0,1]^d$ and the simplex
\begin{equation*}
        \mathcal{X}^{\sort} := \left\{x = (x^1, \cdots, x^d) \in \mathcal{X}\ |\ x^1\geq \cdots \geq x^d\right\}.
    \end{equation*}
Later we partition both domains into smaller $d$-dimensional cubes and simplexes to derive our main results; thus, the only interior cone conditions we need are the following. 

\begin{lemma}\label{computeinteriorconelemma}
    The domain $\mathcal{X} = [0,1]^d$ satisfies the $(\theta_d := \arcsin 1/\sqrt{d}, r_d := 1/2)$-interior cone condition. The domain $ \mathcal{X}^{\sort}$
    satisfies the $(\theta^{\sort}_d := \arcsin 1/d^{3/2}, r^{\sort}_d := 1/(2d+2))$-interior cone condition.
\end{lemma}

A related result is \cite[Proposition 11.26]{wendland2004scattered}, which gives an interior cone condition for a general domain that is \emph{star-shaped with respect to a ball}. However, we note that a direct application of \cite[Proposition 11.26]{wendland2004scattered} yields a $\theta_d$ that is roughly half of the $\theta_d$ from Lemma \ref{computeinteriorconelemma}. Our proof of Lemma \ref{computeinteriorconelemma} leads to a sharper result by examining the vectors from the vertices to the \emph{centers of gravity} and the projections to each of the domain's faces. The detail is given in the appendix.


\paragraph{An overview of the main results.}We will now analyze the errors of approximating $f\in \mathcal{H}$ with $\hat{f}_n := \mathcal{K}(.,X)\mathcal{K}(X,X)^{-1}y$ and approximating $f\in \mathcal{H}^{perm}$ with $\hat{f}^{\sort}_n := \mathcal{K}^{\sort}(.,X)\mathcal{K}^{\sort}(X,X)^{-1}y$, point-wise (Theorem \ref{prop:hpermembeddedinhsortlocal}) and in $L^2(\mathcal{X};\mathbb{P})$-norm (Theorem \ref{thm:hpermembeddedinhsort}). We partition $\mathcal{X} = [0,1]^d$ into $q^d$ equal subcubes of side length $l$ and denote the subcube $(q_1l,\cdots,q_dl) + [0,l]^d \subset \mathcal{X}$ by $\Omega_{q_1,\cdots,q_d}$ for $(q_1,\cdots,q_d) \in I := \{0,\cdots,q-1\}^d$. The choice of $q$ will depend on the underlying fill distance, $h_{X,\mathcal{X}}$ versus $h_{\sort X,\mathcal{X}^{\sort}}$. An approximation of $f \in \mathcal{H}^{perm}$ at any $x$ far from the partial diagonal of $\mathcal{X}^{\sort}$ would enjoy a convergence rate in terms of $h_{\sort X, \mathcal{X}^{\sort}}$ under the sorted kernel interpolation, compared to $h_{X, \mathcal{X}}$ under the unsorted kernel interpolation. Later in Proposition \ref{existencestatementproposition}, we show that $h_{\sort X, \mathcal{X}^{\sort}} \leq h_{X, \mathcal{X}}$ via the \emph{rearrangement principle}, which illustrates the advantage of the sorted kernel approximation. For $x$ near any of the simplex's partial diagonals, we will see from (\ref{sortedapproxlocal}) in Theorem \ref{prop:hpermembeddedinhsortlocal} that, the error bound is $(2d^2)^{\frac{\nu}{2}}$ times of that associated with $x$ far from the partial diagonal of $\mathcal{X}^{\sort}$.  
However, as we show in Theorem \ref{thm:hpermembeddedinhsort}, for a dataset with large $n$ and a small fill distance relative to the dimension $d$, the measure of the set of $x$ with slower point-wise convergence becomes negligible. In this case, an approximation of $f\in\mathcal{H}^{perm}$ would enjoy a faster $L^2(\mathcal{X};\mathbb{P})$-convergence rate in terms of $h_{\sort X, \mathcal{X}^{\sort}}$ under the sorted kernel interpolation, compared to $h_{X, \mathcal{X}}$ under the unsorted kernel interpolation. These results provide a theoretical justification of the canonicalized kernel approximation, despite the poor analytical properties. 

\bigskip

To facilitate the presentation, we introduce a few additional definitions in the following. Given any $\mathcal{K} \in C^\nu(\mathcal{X}\times\mathcal{X})$, we define
\begin{equation*}
    C_{\mathcal{K}, \nu} := \max_{x\in \mathcal{X}}C_{\mathcal{K}, \nu, x, D} = \max_{x_1, x_2 \in \mathcal{X}}\max_{|\alpha|+|\beta|=\nu}\frac{1}{\alpha!\beta!}\left|\partial_1^\alpha\partial_2^\beta\mathcal{K}(x_1,x_2)\right|,
\end{equation*}
and
\begin{equation*}
    \widetilde{C}_{\mathcal{K},\nu,d} := 8{\nu + 2d\choose 2d}C_{\mathcal{K},\nu}\cdot \left(16\nu^2d\right)^{\nu}.
\end{equation*}

\begin{theorem}[Point-wise approximation error]\label{prop:hpermembeddedinhsortlocal} Consider an RKHS $\mathcal{H}$ subject to Assumption \ref{pd before sort} and a design sequence $X = \{x_i\}_{i=1}^{n}$. (i) If
    \begin{equation}
    \left\lfloor \left(8\nu^2(\sqrt{d}+1)h_{X,\mathcal{X}}\right)^{-1} \right\rfloor > 1, \label{cone condition cube}
    \end{equation}
    then, for all $x\in \mathcal{X}$ and $f\in \mathcal{H}$, 
    \begin{equation}\label{standardapproxlocal}
        \left|f(x) - \hat{f}_n(x)\right| \leq \|f\|_{\mathcal{H}}\cdot \sqrt{\widetilde{C}_{\mathcal{K},\nu,d}\cdot (h_{X,\mathcal{X}})^\nu}.
    \end{equation}
    (ii) Furthermore, suppose Assumption \ref{PD on data} is also satisfied.
    If 
    \begin{equation} 
    \left\lfloor \left(8\nu^2(d+1)(d^{3/2}+1)h_{\sort X, \mathcal{X}^{\sort}}\right)^{-1} \right\rfloor > 1, \label{cone condition simplex}
    \end{equation}

   then, for all $x \in \mathcal{X}$ and $f \in \mathcal{H}^{perm}$,
    \begin{equation}\label{sortedapproxlocal}
        \left|f(x) - \hat{f}^{\sort}_n(x)\right| \leq \begin{cases}
            \displaystyle\|f\|_{\mathcal{H}}\cdot \sqrt{\widetilde{C}_{\mathcal{K},\nu, d}\cdot \left(h_{\sort X, \mathcal{X}^{\sort}}\right)^{\nu}}, &\ \text{$x\in \Omega_{q_1,\cdots,q_d}; (q_1,\cdots,q_d)\in I_0$}\\
            \displaystyle\|f\|_{\mathcal{H}}\cdot \sqrt{\widetilde{C}_{\mathcal{K},\nu, d}\cdot \left(2d^2h_{\sort X, \mathcal{X}^{\sort}}\right)^{\nu}}, &\ \text{$x\in \Omega_{q_1,\cdots,q_d}; (q_1,\cdots,q_d)\in I_\partial$}
        \end{cases}
    \end{equation}
     where $\Omega_{q_1,\cdots,q_d} := (q_1l,\cdots,q_dl) + [0,l]^d$ for $(q_1,\cdots,q_d) \in I := \{0,\cdots,q-1\}^d$ are subcubes of $\mathcal{X}$ of side length $l \geq 8\nu^2(\sqrt{d}+1)h_{\sort X,\mathcal{X}^{\sort}}=:l^*$, $q := \lfloor 1/l \rfloor$. We decompose $I := I_0\coprod I_{\partial}$ such that $I_0$ consists of $(q_1,\cdots,q_d)$ with all coordinates distinct from each other, and $I_{\partial}$ consists of the rest of the indices. 
\end{theorem} 

\begin{proof}
    To prove (i), we choose $l \geq 8\nu^2\left(1 + \frac{1}{\sin\theta_d}\right)h_{X, \mathcal{X}}=:l^*$ and define $q := \left\lfloor 1/l\right\rfloor$. We divide $\mathcal{X} = [0,1]^d$ into $q^d$ subcubes of equal side length $l$ and denote the subcube $(q_1l,\cdots,q_dl) + [0,l]^d \subset \mathcal{X}$ by $\Omega_{q_1,\cdots,q_d}$ for $(q_1,\cdots,q_d) \in I := \{0,\cdots,q-1\}^d$. 
    By (\ref{cone condition cube}), each $\Omega_{q_1,\cdots,q_d}\subset \mathcal{X}$, and by Lemma \ref{computeinteriorconelemma}, each $\Omega_{q_1,\cdots,q_d}$ satisfies the $(\theta_d, r_d)$-interior cone condition, where $\sin\theta_d = 1/\sqrt{d}$ and $r_d = \frac{1}{2}l \geq 4\nu^2\left(1 + \frac{1}{\sin \theta_d}\right)h_{X, \mathcal{X}} = 4\nu^2(\sqrt{d}+1)h_{X, \mathcal{X}}$. For any $x \in \mathcal{X}$, we have $x \in \Omega_{q_1,\cdots,q_d}$ for some $(q_1,\cdots,q_d) \in \{0,\cdots,q-1\}^d$. Hence, by setting $\Omega_x = \Omega_{q_1,\cdots,q_d}$, the condition needed for the application of Theorem \ref{thm:localapproxerror} is satisfied with $D := \diam \Omega_x = \sqrt{d}l$. Then we have
    \begin{multline*}
        \left|f(x) - \hat{f}_n(x)\right|^2 \leq \|f\|^2_{\mathcal{H}}\cdot 8{\nu + 2d\choose 2d}C_{\mathcal{K},\nu, x, D}\cdot \left(\sqrt{d}\cdot 8\nu^2(\sqrt{d}+1)h_{X, \mathcal{X}}\right)^{\nu}\\
        \leq \|f\|^2_{\mathcal{H}}\cdot 8{\nu + 2d\choose 2d}C_{\mathcal{K},\nu}\cdot \left(16\nu^2dh_{X, \mathcal{X}}\right)^{\nu}
    \end{multline*}
   for all $x\in \mathcal{X}$.

   \bigskip

    Now let us turn to part (ii). We apply the argument above for $\hat{f}_n^{\sort}$ 
and $f \in \mathcal{H}^{perm}$, where $f = f\circ \sort$. We restrict the fundamental domain to $\mathcal{X}^{\sort}$ and consider a new sequence of design points $\sort X := \{\sort x_i\}_{i=1}^n\subset \mathcal{X}^{\sort}$ with fill distance $h_{\sort X, \mathcal{X}^{\sort}}$. With this sequence of design points, we have $\hat{f}_n = \sum_{i=1}^n\hat{\pi}_i\mathcal{K}(\sort x_i,.)$ where $\hat{\pi} = \mathcal{K}(\sort X, \sort X)^{-1}y = \mathcal{K}^{\sort}(X,X)^{-1}y = \hat{\pi}^{\sort}$, and therefore,
    \begin{equation}
        \hat{f}_n\circ \sort = \sum_{i=1}^n\hat{\pi}^{\sort}_i\mathcal{K}(\sort x_i, \sort .) = \sum_{i=1}^n\hat{\pi}^{\sort}_i\mathcal{K}^{\sort}(x_i,.) = \hat{f}^{\sort}_n. \label{eq:f_hat_sort}
    \end{equation}  
    Let us divide $\mathcal{X} = [0,1]^d$ into $q^d$ subcubes $\Omega_{q_1,\cdots,q_d}$ of equal side length $l$, similarly to part (i), but with $h_{\sort X, \mathcal{X}^{\sort}}$ instead of $h_{X,\mathcal{X}}$ (see part (ii) of the theorem's statement). For any $x \in \mathcal{X}^{\sort}$, if $x \in \Omega_{q_1,\cdots,q_d}$ where $q_1 > \cdots > q_d$ are all distinct from each other, we choose $\Omega_x = \Omega_{q_1,\cdots,q_d}$. On the other hand, if $x\in \Omega_{q_1,\cdots,q_d}$ with some repeating $q_1\geq \cdots \geq q_d$, then we set $\Omega_x = ((q_1l,\cdots,q_dl) + [0,s\cdot l']^d)\cap\mathcal{X}^{\sort}$ where $l' := 8\nu^2(d+1)(d^{3/2}+1)h_{\sort X, \mathcal{X}^{\sort}}$ and $s\in \{-1,+1\}$ is chosen appropriately so that $(q_1l,\cdots,q_dl) + [0,s\cdot l']^d\subset \mathcal{X}$. Note that such a choice of $s$ is possible by the condition (\ref{cone condition simplex}). For example, if $q_1l$ is too close to $1$, then we choose $s = -1$. If $q_1 = \cdots = q_d$, then $\Omega_x$ is a simplex of side length $l'$, and by Lemma \ref{computeinteriorconelemma}, $\Omega_x$ satisfies the interior cone condition with $\sin\theta^{\sort}_d = 1/d^{3/2}$ and $r^{\sort}_d = \frac{1}{2(d+1)}l' = 4\nu^2\left(1 + \frac{1}{\sin \theta^{\sort}_d}\right)h_{\sort X,\mathcal{X}^{\sort}} = 4\nu^2(d^{3/2}+1)h_{\sort X,\mathcal{X}^{\sort}}$. Other cases where $q_1,\cdots,q_d$ are not all distinct but not all identical are less restrictive, and therefore $\Omega_x$ also satisfies the $(\theta^{\sort}_d, r^{\sort}_d)$-interior cone condition in these cases. Therefore, the condition needed for the application of Theorem \ref{thm:localapproxerror} on $\mathcal{X}^{\sort}$ is satisfied with $D := \diam \Omega_x = \sqrt{d}l$ if $q_1,\cdots,q_d$ are all distinct, and $D = \sqrt{d}l'$ otherwise. So we have for all $x \in \mathcal{X}^{\sort}$,
    \begin{equation*}
        \left|f(x) - \hat{f}_n(x)\right|^2 \leq \begin{cases}
            \|f\|^2_{\mathcal{H}}\cdot 8{\nu + 2d\choose 2d}C_{\mathcal{K},\nu}\cdot \left(16\nu^2dh_{\sort X, \mathcal{X}^{\sort}}\right)^{\nu}, &\ \text{$x\in \Omega_{q_1,\cdots,q_d}; (q_1,\cdots,q_d)\in I_0$}\\
            \|f\|^2_{\mathcal{H}}\cdot 8{\nu + 2d\choose 2d}C_{\mathcal{K},\nu}\cdot \left(32\nu^2d^3h_{\sort X, \mathcal{X}^{\sort}}\right)^{\nu}, &\ \text{$x\in \Omega_{q_1,\cdots,q_d}; (q_1,\cdots,q_d)\in I_\partial$}
        \end{cases}.
    \end{equation*}
    For an arbitrary $x\in \mathcal{X}$, we have $\sort x\in \mathcal{X}^{\sort}$. Hence, the result follows by noting that $\hat{f}^{\sort}_n = \hat{f}_n\circ \sort$, and since $f\in \mathcal{H}^{perm}$, we also have $f = f\circ\sort$.
\end{proof}




Next, we present the bounds on the approximation errors in the $L^2(\mathcal{X};\mathbb{P})$-norm. 

\begin{theorem}[Approximation error in $L^2(\mathcal{X};\mathbb{P})$-norm]\label{thm:hpermembeddedinhsort}
Consider an RKHS $\mathcal{H}$ subject to Assumption \ref{pd before sort} and a design sequence $X = \{x_i\}_{i=1}^{n}$. Suppose that the domain $\mathcal{X}=[0,1]^d$ is equipped with probability measure $\mathbb{P}$ that has a continuous density bounded in $[\underline{\rho}, \bar{\rho}] \subset \mathbb{R}_{>0}$. (i) Then

    \begin{equation}\label{standardapprox}
        \left\|f - \hat{f}_n\right\|^2_{L^2(\mathcal{X}, \mathbb{P})} \leq \|f\|^2_{\mathcal{H}}\cdot \widetilde{C}_{\mathcal{K},\nu,d}\cdot \left(h_{X, \mathcal{X}}\right)^{\nu}
    \end{equation}

for all sufficiently large $n$ such that (\ref{cone condition cube}) holds. (ii) Additionally, suppose that Assumption \ref{PD on data} holds. Then for any arbitrary $\alpha > 1$, we have

\begin{equation}\label{sortedapproacheszero}
        \left\|f - \hat{f}^{\sort}_n\right\|^2_{L^2(\mathcal{X}, \mathbb{P})} \leq \alpha \cdot\|f\|^2_{\mathcal{H}}\cdot \widetilde{C}_{\mathcal{K},\nu,d}\cdot\left(h_{\sort X, \mathcal{X}^{\sort}}\right)^{\nu}
    \end{equation}

when $P_{\nu,d}h_{\sort X,\mathcal{X}^{\sort}}=o(1)$ as $n\rightarrow \infty$, where $P_{\nu,d} := 8\bar{\rho}\cdot 2^{\nu}\nu^2d^{2\nu+5/2}$.
\end{theorem}




\begin{proof} 
    Both results are obtained by integrating the corresponding parts of the local results in Theorem \ref{prop:hpermembeddedinhsortlocal}. For part (i), we have from (\ref{standardapproxlocal}) that
    \begin{equation*}
        \left\|f - \hat{f}_n\right\|_{L^2(\mathcal{X}, \mathbb{P})}^2 = \int_{\mathcal{X}}\left|f(x) - \hat{f}_n(x)\right|^2\mathbb{P}(dx) \leq \|f\|^2_{\mathcal{H}}\cdot \widetilde{C}_{\mathcal{K},\nu,d}\cdot \left(h_{X, \mathcal{X}}\right)^{\nu}.
    \end{equation*}

    For part (ii), let $q := \lfloor 1/l^* \rfloor$ where $l^*$ is given in Theorem \ref{prop:hpermembeddedinhsortlocal}. We note that $q \geq \left(16\nu^2\sqrt{d}h_{\sort X, \mathcal{X}^{\sort}}\right)^{-1}$. Out of the $q^d$ subcubes, the number of subcubes $\Omega_{q_1,\cdots,q_d}$ with $q_1,\cdots,q_d$ all distinct from each other is $\frac{q!}{(q-d)!}$. Moreover, recall we assume that the probability density is bounded above by $\bar{\rho}$. It follows that $\bar{\rho}\cdot\left(1 - \frac{q!}{q^d(q-d)!}\right)$ gives an upper bound for the measure of $\coprod_{(q_1,\cdots,q_d) \in I_\partial}\Omega_{q_1,\cdots,q_d}$. Therefore, we have the following estimate from (\ref{sortedapproxlocal}):
    \begin{multline*}
        \left\|f - \hat{f}^{\sort}_n\right\|_{L^2(\mathcal{X}, \mathbb{P})}^2 = \int_{\mathcal{X}}\left|f(\sort x) - \hat{f}_n(\sort x)\right|^2\mathbb{P}(dx)\\
        \leq \|f\|^2_{\mathcal{H}}\cdot \widetilde{C}_{\mathcal{K}, \nu, d}\cdot \left(1 + \bar{\rho}\cdot\left(1 - \frac{q!}{q^d(q-d)!}\right)\cdot (2d^2)^\nu\right)\left(h_{\sort X, \mathcal{X}^{\sort}}\right)^{\nu}.
    \end{multline*}
    To finish the proof, we note that
        \begin{equation*}
        1-\frac{q!}{q^d(q-d)!} = 1 - \prod_{k=0}^{d-1}\left(1 - \frac{k}{q}\right) \leq \sum_{k=0}^{d-1}\frac{k}{q} = \frac{d(d-1)}{2q} \leq 8\nu^2d^{3/2}(d-1)h_{\sort X, \mathcal{X}^{\sort}},
    \end{equation*}
    which gives \begin{equation}\label{sortedapprox}
        \left\|f - \hat{f}^{\sort}_n\right\|^2_{L^2(\mathcal{X}, \mathbb{P})} \leq \|f\|^2_{\mathcal{H}}\cdot \widetilde{C}_{\mathcal{K},\nu,d}\cdot \left(1 + P_{\nu,d}h_{\sort X, \mathcal{X}^{\sort}}\right)\cdot \left(h_{\sort X, \mathcal{X}^{\sort}}\right)^{\nu}
    \end{equation}
    where $P_{\nu,d} := 8\bar{\rho}\cdot 2^{\nu}\nu^2d^{2\nu+5/2}$. Under $2^\nu \nu^2 d^{2\nu+5/2}h_{\sort X_n,\mathcal{X}^{\sort}}=o(1)$, we have (\ref{sortedapproacheszero}).
\end{proof}

\bigskip
The fill distance plays a key role in the bound for the approximation error and depends on the design of sample points. The next result compares $h_{\sort X, \mathcal{X}^{\sort}}$ with $h_{X, \mathcal{X}}$.

\begin{proposition}\label{existencestatementproposition}
     For any given design sequence $X = \{x_i\}_{i=1}^n$, we have $h_{\sort X, \mathcal{X}^{\sort}}\leq h_{X, \mathcal{X}}$. In addition, there exists a design sequence $X = \{x_i\}_{i=1}^n$ such that $h_{\sort X, \mathcal{X}^{\sort}}\leq \frac{1}{(d!)^{1/d}}h_{X, \mathcal{X}}$.
\end{proposition}

\begin{proof}
For the first part, note that given any $x_1, x_2 \in \mathcal{X} = [0,1]^d$, we have 
\begin{multline*}
    \|\sort x_1 - \sort x_2\|^2_2 = \|\sort x_1\|^2_2 + \|\sort x_2\|^2_2 - 2(\sort x_1)\cdot (\sort x_2)\\
    = \|x_1\|^2_2 + \|x_2\|^2_2 - 2(\sort x_1)\cdot (\sort x_2)\\
    \leq \|x_1\|^2_2 + \|x_2\|^2_2 - 2x_1\cdot x_2 = \|x_1-x_2\|^2_2.
\end{multline*}
The second equality follows from the fact that sorting (or any coordinate permutation) is an isometry. The inequality follows from the fact that $x_1\cdot x_2 \geq 0$ is larger when the coordinates of both $x_1$ and $x_2$ are sorted under the same ordering as the result of the rearrangement inequality. Therefore, $h_{\sort X, \mathcal{X}^{\sort}} \leq h_{X, \mathcal{X}}$ for any given design sequence $X$. 

    \bigskip

    For the second part, we can choose $X = \{x_i\}_{i=1}^{n} \subset \mathcal{X}$ to be an $\varepsilon$-covering of $\mathcal{X}^{\sort}(\subset \mathcal{X})$ with the minimal cardinality. We have $n = N(\varepsilon, \mathcal{X}^{\sort})$, the $\varepsilon$-covering number of $\mathcal{X}^{\sort}$. In particular, we can assume that $X = \sort X \subset \mathcal{X}^{\sort}$. Then $h_{\sort X, \mathcal{X}^{\sort}} \leq \varepsilon$ while $h_{X, \mathcal{X}} \geq 1$, since $X\cap (\mathcal{X}\setminus \mathcal{X}^{\sort})=\emptyset$. Since $\mathcal{X}^{\sort}$ is convex and $B(x,\varepsilon)\subset \mathcal{X}^{\sort}$ for all sufficiently small $\varepsilon > 0$, we obtain the following bound: $N(\varepsilon, \mathcal{X}^{\sort}) \leq \frac{1}{d!\omega_d}\left(\frac{3}{\varepsilon}\right)^d$. This gives the upper bound for the fill distance: $h_{\sort X,\mathcal{X}^{\sort}} \leq \varepsilon \leq 3(d!\omega_dn)^{-1/d}$.
\end{proof}

\textbf{Rearrangement principle.} The first part of Proposition \ref{existencestatementproposition} highlights the potential improvement of using $\mathcal{K}(\sort ., \sort .)$ in approximating a permutation invariant function over using the unsorted kernel $\mathcal{K}(.,.)$ and is rooted in the \emph{rearrangement principle}: if $w^1\geq\cdots\geq w^d$ and $z^1\geq\cdots\geq z^d$, then $w^1z^1+\cdots+w^dz^d \geq w^1z^{\sigma(1)}+\cdots+w^dz^{\sigma(d)}$ for every permutation $\sigma \in S_d$. 

\bigskip

The second part of Proposition \ref{existencestatementproposition} suggests that, under the setup of Theorem \ref{thm:hpermembeddedinhsort}(ii), there exists a design sequence $X = \{x_i\}_{i=1}^n$ such that for all $f \in \mathcal{H}^{perm}$,
    \begin{equation}\label{sortedapproxdhapproacheszerogrid}
        \left\|f - \hat{f}^{\sort}_n\right\|^2_{L^2(\mathcal{X}, \mathbb{P})} \leq \frac{3^\nu\alpha \cdot\|f\|^2_{\mathcal{H}}\cdot \widetilde{C}_{\mathcal{K},\nu,d}}{\left(d!\omega_d\right)^{\nu/d}n^{\nu/d}},
\end{equation}
while 
\begin{equation}\label{standardapproxdhapproacheszerogrid}
      \left\|f - \hat{f}_n\right\|^2_{L^2(\mathcal{X}, \mathbb{P})} \leq \frac{3^\nu\cdot\|f\|^2_{\mathcal{H}}\cdot \widetilde{C}_{\mathcal{K},\nu,d}}{\left(\omega_d\right)^{\nu/d}n^{\nu/d}}.
\end{equation}

The proof underlying the second part of Proposition \ref{existencestatementproposition} chooses a design that targets on $\mathcal{X}^{\sort}$. In this case, one saves resources in interpolating $f\in \mathcal{H}^{perm}$ as well as data collection. This observation prompts an interesting future research question on how one can choose a design sequence in $\mathcal{X}$ to fully leverage the sorting technique. 

\bigskip

\textbf{Extension to finite group actions.}
We can consider an extension of Theorem \ref{thm:hpermembeddedinhsort} for a general finite group $G$ acting on a compact domain $\mathcal{X}\subset \mathbb{R}^d$. We shall assume that the indicator function of $\mathcal{X}/G\hookrightarrow \mathcal{X}$ is Riemann-integrable. Let us cover $\mathcal{X}/G$ with a collection $\{\Omega_q\}_{q \in I}$ of $d$-dimensional cubes of side length $l := 8\nu^2(\sqrt{d}+1)h_{\pi X, \mathcal{X}/G}$, where $I$ is some index set. Let us decompose $I = I_0\coprod I_\partial$, where $I_\partial$ consists of indices $q$ such that $\Omega_q \cap \partial (\mathcal{X}/G) \neq \emptyset$. Then (\ref{sortedapprox}) can be generalized for the $\mathcal{H}^\pi$ minimal norm interpolant $\hat{f}^\pi_n$ as follows:
    \begin{equation*}
        \left\|f - \hat{f}^\pi_n\right\|^2_{L^2(\mathcal{X}, \mathbb{P})} \leq \|f\|^2_{\mathcal{H}}\cdot \widetilde{C}_{\mathcal{K}, \nu, d}\cdot \left(\int_{\coprod_{q\in I_0}\Omega_q}1\cdot \mathbb{P}(dx) + \int_{\coprod_{q \in I_\partial}\Omega_q}\left(\frac{D_x/r^\pi_{d,x}}{d\sin\theta^\pi_{d,x}}\right)^\nu \mathbb{P}(dx) \right)\left(h_{\pi X, \mathcal{X}/G}\right)^{\nu}
    \end{equation*}
    where $x\in\Omega_x \subset \mathcal{X}/G$ for some $\Omega_x$ satisfying the $(r^\pi_{d,x}, \theta^\pi_{d,x})$-interior cone condition, and $D_x := \diam \Omega_x$ for any $x \in \Omega_q$, $q \in I_\partial$. Note that $r^\pi_{d,x}/D_x$ is essentially the interior cone radius if $\Omega_x$ is rescaled to a unit diameter. Therefore, $D_x/(r^\pi_{d,x}\sin \theta^\pi_{d,x})$ encodes the geometric information of $\Omega_x$ and hence, of $\Omega_q\cap\mathcal{X}/G$ for $q \in I_\partial$. By Riemann integrability, $\partial(\mathcal{X}/G)$ has measure zero, and therefore, the second integral above would approach $0$ while the first integral approaches $1$, asymptotically in $n$.\footnote{In particular, if $G\leq S_d$ is any subgroup acting on $\mathcal{X} = [0,1]^d$ via coordinate permutation, then we have $\mathcal{X}^{\sort} \subseteq \mathcal{X}/G \hookrightarrow \mathcal{X}$. In this case, the interior cone angle and radius we derive in Lemma \ref{computeinteriorconelemma} remains valid but might be conservative.}

    \bigskip

Finally, Theorem \ref{thm:hpermembeddedinhsort} provides an embedding of $\mathcal{H}^{perm}$ into $\overline{\mathcal{H}}^{\sort}$. This theoretical insight relates the two RKHSs $\mathcal{H}^{perm}$ and $\mathcal{H}^{\sort}$. More practically, this result shows that $\mathcal{K}^{\sort}$ is capable of reproducing any function in $\mathcal{H}^{perm}$. Therefore, when performing interpolation with the ground truth function known to be in $\mathcal{H}^{perm}$, we may replace $\mathcal{K}^{perm}$ with $\mathcal{K}^{\sort}$. While computing $\mathcal{K}^{perm}$ from $\mathcal{K}$ involves averaging $\mathcal{K}$ over $(d!)^2$ permutations, with sorting, one simply takes the kernel function associated with the original RKHS and sorts the inputs. 

\begin{corollary}\label{embeddingcorollary}
    Consider the same setup as in Theorem \ref{thm:hpermembeddedinhsort}(ii). Given a positive definite kernel $\mathcal{K} \in C^\nu(\mathcal{X}\times \mathcal{X})$, we have an embedding $\mathcal{H}^{perm}\subset \overline{\mathcal{H}}^{\sort}$ as subspaces of $L^2(\mathcal{X},\mathbb{P})$.
\end{corollary}
\begin{proof}
    By properly choosing a sequence of sample points, 
    we can ensures that $h_{\sort X, \mathcal{X}^{\sort}} \rightarrow 0$ as $n\rightarrow \infty$. Given $f \in \mathcal{H}^{perm}$, we have from Theorem \ref{thm:hpermembeddedinhsort}(ii) that a sequence $\left\{\hat{f}^{\sort}_n\right\}_{n=1}^\infty \subset\mathcal{H}^{\sort}$ converges to $f$ with respect to the $L^2(\mathcal{X},\mathbb{P})$-norm, which means $f \in \overline{\mathcal{H}}^{\sort}$. In other words, $\mathcal{H}^{perm} \subset \overline{\mathcal{H}}^{\sort}$.
\end{proof}

\subsection{Numerical examples}

We illustrate the sorting trick with a simple kernel interpolation example. The target permutation invariant function is the sum of sinusoids:
\begin{equation*}
f(x)=\sum_{a=1}^d \sin(2\pi x^a),\qquad x\in[0,1]^d.
\end{equation*}
Two commonly used kernels are considered: a one‑hidden‑layer neural network kernel and a Gaussian kernel,
\begin{equation*}
\mathcal{K}_{\mathrm{1LNN}}(w,z)=\frac{2}{\pi}\arcsin\left(\frac{2\langle \tilde w,\tilde z\rangle}{\sqrt{(1+2\|\tilde w\|^2)(1+2\|\tilde z\|^2)}}\right), \quad \mathcal{K}_{\mathrm{Gauss}}(w,z)=\exp\left(-10\|w-z\|^2\right),   
\end{equation*}
 
where $\tilde{w}=(1,w)$ and $\tilde{z} = (1,z)$. For each dimension $d\in\{3,6,9,12\}$, we draw $n=200$ points independently from the uniform distribution on $[0,1]^d$. We reserve $100$ points for training and the remaining $100$ points for numerically estimating $\|f-\hat{f}_n\|^2_{L^2(\mathcal{X};\mathbb{P})}$ and $\|f-\hat{f}^{\sort}_n\|^2_{L^2(\mathcal{X};\mathbb{P})}$. The results are reported in Table \ref{tab:simulation}.

\begin{table}[ht]
\centering
\caption{Mean squared prediction error of the sum of sinusoids target function}
\label{tab:simulation}
\begin{tabular}{cccccc}
\toprule
$d$ & $\mathcal{K}_{\mathrm{1LNN}}$ & $\mathcal{K}^{\sort}_{\mathrm{1LNN}}$ & $\mathcal{K}_{\mathrm{Gauss}}$ & $\mathcal{K}^{\sort}_{\mathrm{Gauss}}$ \\
\midrule
3  & 0.024 & 0.006 & 0.020 & 0.001 \\
6  & 0.972  & 0.0491  & 1.519  & 0.106   \\
9  & 3.060  & 0.111   & 4.487  & 0.321   \\
12 & 3.991  & 0.089  & 6.043  & 0.498   \\
\bottomrule
\end{tabular}
\end{table}

In all cases, the sorted kernel outperforms the unsorted counterpart, and the gap appears to grow with the dimension. This observation is consistent with our theoretical insight that sorting reduces the fill distance and confines the approximation to the smaller fundamental domain.  
 
\subsection{Probabilistic bounds}

The next proposition bounds $h_{X,\mathcal{X}}$ and $h_{\sort X,\mathcal{X}^{\sort}}$ when $X = \{x_i\}_{i=1}^n$ are independently and identically drawn from a distribution $\mathbb{P}$ with a continuous density bounded below by $\underline{\rho} > 0$.

\begin{proposition}\label{probabilityboundsforhlemma}
    Consider a design sequence $X = \{x_i\}_{i=1}^n$ independently drawn from a distribution $\mathbb{P}$ with a continuous density bounded below by $\underline{\rho} > 0$. Then we have
    \begin{equation}\label{probabilityboundsforh}
    \begin{aligned}
        \mathbb{P}\left[h_{X,\mathcal{X}} > \varepsilon\right] &< \frac{1}{\omega_d}\left(\frac{6}{\varepsilon}\right)^d\cdot \left(1 - \underline{\rho}\omega_d\left(\frac{\varepsilon}{4}\right)^d\right)^n,\\
        \mathbb{P}\left[h_{\sort X,\mathcal{X}^{\sort}} > \varepsilon\right] &< \frac{1}{d!\omega_d}\left(\frac{6}{\varepsilon}\right)^d\cdot \left(1 - \underline{\rho}\omega_d\left(\frac{\varepsilon}{4}\right)^d\right)^n,
    \end{aligned}
    \end{equation}
    for all sufficiently small $\varepsilon > 0$ such that $B(x, \varepsilon)\subset \mathcal{X}$ for some $x \in \mathcal{X}$, and $B(x, \varepsilon)\subset \mathcal{X}^{\sort}$ for some $x \in \mathcal{X}^{\sort}$, respectively.
\end{proposition}
\begin{proof}
    Consider an $\varepsilon/2$-covering with minimal cardinality, i.e., a set $\{v_1,\cdots,v_N\} \subset \mathcal{X}$ such that for any $x \in \mathcal{X}$, there exists $v_j \in \{v_1,\cdots,v_N\}$ such that $x\in B(v_j, \varepsilon/2)$. We have $N = N(\varepsilon/2,\mathcal{X})$, the $\varepsilon/2$-covering number of $\mathcal{X}$. If $B(v_j, \varepsilon/2)\cap X \neq \emptyset$ for all $j=1,\cdots,N$, then $\min_{i=1,\cdots,n}\|x-x_i\|_2 \leq \|x-v_j\|_2 + \min_{x'\in B(v_j, \varepsilon/2)\cap X}\|v_j - x'\|_2\leq \varepsilon/2 + \varepsilon/2 = \varepsilon$ for all $x \in \mathcal{X}$, which means $h_{X, \mathcal{X}} \leq \varepsilon$. It follows that
    \begin{multline*}
        \mathbb{P}[h_{X, \mathcal{X}} > \varepsilon] \leq \mathbb{P}\left[\bigcup_{j=1,\cdots,N}\{B(v_j, \varepsilon/2)\cap X=\emptyset\}\right] \leq \sum_{j=1,\cdots,N}\mathbb{P}\left[B(v_j,\varepsilon/2)\cap X=\emptyset\right]\\
        \leq N(\varepsilon/2, \mathcal{X})\cdot\left(1-\underline{\rho}\min_{x\in \mathcal{X}}|B(x,\varepsilon/2)\cap\mathcal{X}|\right)^n.
    \end{multline*}
    For $\mathcal{X} = [0,1]^d$, we can bound $N(\varepsilon/2, \mathcal{X})$ and $\underline{\rho}\min_{x\in \mathcal{X}}|B(x,\varepsilon/2)\cap\mathcal{X}|$ more explicitly as follows. Let $\{\tilde{v}_1,\cdots,\tilde{v}_M\}\subset \mathcal{X}$ be a maximal $\varepsilon/2$-packing. Then $N \leq M$ and $\coprod_{j=1,\cdots,M}\left(\tilde{v}_j + B(0,\varepsilon/4)\right)\subset \mathcal{X} + B(0,\varepsilon/4) \subset \mathcal{X} + \frac{1}{2}\mathcal{X} \subset \frac{3}{2}\mathcal{X}$. The second last inclusion follows from the assumption $B(x,\varepsilon)\subset \mathcal{X}$ for some $x \in \mathcal{X}$, and the last inclusion follows from the fact that $\mathcal{X}$ is convex. This implies $M\left|B(0,\varepsilon/4)\right| \leq \left|\frac{3}{2}\mathcal{X}\right|$, and therefore,
    \begin{equation*}
        N(\varepsilon/2,\mathcal{X})\leq \frac{\left|\frac{3}{2}\mathcal{X}\right|}{|B(0,\varepsilon/4)|} = \frac{(3/2)^d}{\omega_d(\varepsilon/4)^d} = \frac{1}{\omega_d}\left(\frac{6}{\varepsilon}\right)^d.
    \end{equation*}
    On the other hand, we have $\min_{x\in \mathcal{X}}|B(x,\varepsilon/2)\cap\mathcal{X}| = \frac{1}{2^d}\omega_d(\varepsilon/2)^d = \omega_d(\varepsilon/4)^d$, where the factor $\frac{1}{2^d}$ accounts for the fact that only points in $B(0,\varepsilon/2)$ with positive coordinates are in $\mathcal{X}$.

\bigskip 
    Similarly, we have
    \begin{equation*}
        \mathbb{P}[h_{\sort X, \mathcal{X}^{\sort}} > \varepsilon] \leq N(\varepsilon/2, \mathcal{X}^{\sort})\cdot \left(1 - d!\underline{\rho}\min_{x\in \mathcal{X}^{\sort}}|B(x,\varepsilon/2)\cap\mathcal{X}^{\sort}|\right)^n.
    \end{equation*}
    The factor of $d!$ comes from the fact that if the density of the distribution for $x \sim \mathbb{P}$ is bounded below by $\underline{\rho}$, then the density of the distribution for $\sort x$ is bounded below by $d!\underline{\rho}$. We can bound the expression above more explicitly using
    \begin{equation*}
        N(\varepsilon/2, \mathcal{X}^{\sort}) \leq \frac{\left|\frac{3}{2}\mathcal{X}^{\sort}\right|}{|B(0,\varepsilon/4)|} = \frac{(3/2)^d/d!}{\omega_d(\varepsilon/4)^d} = \frac{1}{d!\omega_d}\left(\frac{6}{\varepsilon}\right)^d,
    \end{equation*}
    since we have assumed that $B(x,\varepsilon) \subset \mathcal{X}^{\sort}$ for some $x\in \mathcal{X}^{\sort}$, and $\mathcal{X}^{\sort}$ is convex. Similarly, we have $\min_{x\in \mathcal{X}^{\sort}}|B(x,\varepsilon/2)\cap\mathcal{X}^{\sort}| = \frac{1}{d!2^d}\omega_d(\varepsilon/2)^d = \frac{1}{d!}\omega_d(\varepsilon/4)^d$, which means $d!\underline{\rho}\min_{x\in \mathcal{X}^{\sort}}|B(x,\varepsilon/2)\cap\mathcal{X}^{\sort}| = \underline{\rho}\omega_d(\varepsilon/4)^d$.
\end{proof}

Combining Theorem \ref{thm:hpermembeddedinhsort} and Proposition \ref{probabilityboundsforhlemma} yields the following result. 

\begin{proposition}\label{probabilityboundproposition}
    Consider an RKHS $\mathcal{H}$ subject to Assumption \ref{pd before sort} and a design sequence $X = \{x_i\}_{i=1}^n$ independently drawn from a distribution $\mathbb{P}$ with a continuous density bounded in $[\underline{\rho}, \bar{\rho}] \subset \mathbb{R}_{>0}$. (i) Suppose $\left\lfloor \left(8\nu^2(\sqrt{d}+1)\varepsilon^{1/\nu}\right)^{-1} \right\rfloor > 1$. Then the $L^2(\mathcal{X},\mathbb{P})$ interpolation error for any $f \in \mathcal{H}$ via $\mathcal{K}$ satisfies
    \begin{multline*}\label{expprobabilityboundforf}
        \mathbb{P}\left[\| f - \hat{f}_n\|^2_{L^2(\mathcal{X},\mathbb{P})} > \varepsilon\right] < \frac{6^d}{\omega_d}\left(\frac{\max\{1,\|f\|^2_{\mathcal{H}}\widetilde{C}_{\mathcal{K},\nu,d}\}}{\varepsilon}\right)^{d/\nu}\\
        \times\left(1- \frac{\underline{\rho}\omega_d}{4^d}\left(\frac{\varepsilon}{\max\{1,\|f\|^2_{\mathcal{H}}\widetilde{C}_{\mathcal{K},\nu,d}\}}\right)^{d/\nu}\right)^n.\numberthis
    \end{multline*}
    (ii) Suppose $\left\lfloor \left(8\nu^2(d+1)(d^{3/2}+1)\varepsilon^{1/\nu}\right)^{-1} \right\rfloor > 1$. Then the $L^2(\mathcal{X}, \mathbb{P})$ interpolation error for any $f \in \mathcal{H}^{perm}$ via $\mathcal{K}^{\sort}$ satisfies
    \begin{multline*}\label{expprobabilityboundforfsort}
        \mathbb{P}\left[\| f - \hat{f}^{\sort}_n\|^2_{L^2(\mathcal{X},\mathbb{P})} > \varepsilon\right] < \frac{1}{d!}\cdot\frac{6^d}{\omega_d}\left(\frac{\max\left\{1,\|f\|^2_{\mathcal{H}}\widetilde{C}_{\mathcal{K},\nu,d}\left(1 + P_{\nu,d}\varepsilon^{1/\nu}\right)\right\}}{\varepsilon}\right)^{d/\nu}\\
        \times\left(1- \frac{\underline{\rho}\omega_d}{4^d}\left(\frac{\varepsilon}{\max\left\{1,\|f\|^2_{\mathcal{H}}\widetilde{C}_{\mathcal{K},\nu,d}\left(1 + P_{\nu,d}\varepsilon^{1/\nu}\right)\right\}}\right)^{d/\nu}\right)^n\numberthis
    \end{multline*}
    where $P_{\nu,d} = 8\bar{\rho}\cdot 2^{\nu}\nu^2d^{2\nu+5/2}$.
\end{proposition}

\begin{proof}
    For part (i), the condition $\left\lfloor \left(8\nu^2(\sqrt{d}+1)\varepsilon^{1/\nu}\right)^{-1} \right\rfloor > 1$ ensures that (\ref{cone condition cube}) is satisfied and we can apply the bound in Theorem \ref{thm:hpermembeddedinhsort} if $h_{X,\mathcal{X}} < \varepsilon^{1/\nu}$. Therefore, 
    \begin{multline*}
        \mathbb{P}\left[\| f - \hat{f}_n\|^2_{L^2(\mathcal{X},\mathbb{P})} > \varepsilon\right] \leq \mathbb{P}\left[\varepsilon^{1/\nu} > h_{X,\mathcal{X}} > \left(\frac{\varepsilon}{\|f\|^2_{\mathcal{H}}\widetilde{C}_{\mathcal{K}, \nu, d}}\right)^{1/\nu}\right] + \mathbb{P}\left[h_{X,\mathcal{X}} > \varepsilon^{1/\nu}\right]\\
        \leq  \mathbb{P}\left[h_{X,\mathcal{X}} > \left(\frac{\varepsilon}{\|f\|^2_{\mathcal{H}}\widetilde{C}_{\mathcal{K}, \nu, d}}\right)^{1/\nu}\right] + \mathbb{P}\left[h_{X,\mathcal{X}} > \varepsilon^{1/\nu}\right]\\
        \leq 2\cdot \mathbb{P}\left[h_{X,\mathcal{X}} > \left(\frac{\varepsilon}{\max\{1,\|f\|^2_{\mathcal{H}}\widetilde{C}_{\mathcal{K}, \nu, d}\}}\right)^{1/\nu}\right].
    \end{multline*}
    The condition $\left\lfloor \left(8\nu^2(\sqrt{d}+1)\varepsilon^{1/\nu}\right)^{-1} \right\rfloor > 1$ also ensures that $\mathcal{X} = [0,1]^d$ contains a subcube of side length $l_\varepsilon := 8\nu^2(\sqrt{d}+1)\varepsilon^{1/\nu}$. Such a subcube satisfies the cone condition with $r_d = \frac{1}{2}l_\varepsilon = 4\nu^2\left(1 + \frac{1}{\sin\theta_d}\right)\varepsilon^{1/\nu}$, and hence contains a ball of radius $\varepsilon^{1/\nu}$ (see \cite[Lemma 3.7]{wendland2004scattered}). As a result, there exists $x \in \mathcal{X}$ such that $B\left(x,(\varepsilon/\max\{1,\|f\|^2_{\mathcal{H}}\widetilde{C}_{\mathcal{K}, \nu, d}\})^{1/\nu}\right) \subset B(x,\varepsilon^{1/\nu})\subset \mathcal{X}$, where the first inclusion comes from the fact that $\max\{1,\|f\|^2_{\mathcal{H}}\widetilde{C}_{\mathcal{K}, \nu, d}\} \geq 1$. This allows us to apply Lemma \ref{probabilityboundsforhlemma} for the tail bound of $h_{X,\mathcal{X}}$, from which the result follows.

\bigskip 
    For part (ii), the condition $\left\lfloor \left(8\nu^2(d+1)(d^{3/2}+1)\varepsilon^{1/\nu}\right)^{-1} \right\rfloor > 1$ ensures that (\ref{cone condition simplex}) is satisfied and we can apply the bound in Theorem \ref{thm:hpermembeddedinhsort} if $h_{\sort X,\mathcal{X}^{\sort}} < \varepsilon^{1/\nu}$. From (\ref{sortedapprox}), it follows that $\| f - \hat{f}^{\sort}_n\|^2_{L^2(\mathcal{X},\mathbb{P})} > \varepsilon$ if
    \begin{equation*}
        \frac{\varepsilon}{\|f\|^2_{\mathcal{H}}\cdot \widetilde{C}_{\mathcal{K},\nu,d}} < \left(1 + P_{\nu,d} h_{\sort X, \mathcal{X}^{\sort}}\right)\left(h_{\sort X, \mathcal{X}^{\sort}}\right)^{\nu} < \left(1 + P_{\nu,d}
        \varepsilon^{1/\nu}\right)\left(h_{\sort X, \mathcal{X}^{\sort}}\right)^{\nu},
    \end{equation*}
    where we have used $h_{\sort X, \mathcal{X}^{\sort}} < \varepsilon^{1/\nu}$ in the second inequality. Therefore, we have
    \begin{multline*}
        \mathbb{P}\left[\| f - \hat{f}^{\sort}_n\|^2_{L^2(\mathcal{X},\mathbb{P})} > \varepsilon\right] \leq \mathbb{P}\left[\varepsilon^{1/\nu} > h_{\sort X,\mathcal{X}^{\sort}} > \left(\frac{\varepsilon}{\|f\|^2_{\mathcal{H}}\widetilde{C}_{\mathcal{K}, \nu, d}\left(1 + P_{\nu,d}\varepsilon^{1/\nu}\right)}\right)^{1/\nu}\right]\\
        + \mathbb{P}\left[h_{\sort X,\mathcal{X}^{\sort}} > \varepsilon^{1/\nu}\right] \leq 2\cdot \mathbb{P}\left[h_{\sort X,\mathcal{X}^{\sort}} > \left(\frac{\varepsilon}{\max\left\{1,\|f\|^2_{\mathcal{H}}\widetilde{C}_{\mathcal{K}, \nu, d}\left(1 + P_{\nu,d}\varepsilon^{1/\nu}\right)\right\}}\right)^{1/\nu}\right].
    \end{multline*}
    The condition $\left\lfloor \left(8\nu^2(d+1)(d^{3/2}+1)\varepsilon^{1/\nu}\right)^{-1} \right\rfloor > 1$ also ensures that $\mathcal{X}^{\sort}$ contains a simplex of side length $l'_\varepsilon := 8\nu^2(d+1)(d^{3/2}+1)\varepsilon^{1/\nu}$, which satisfies the cone condition with $r^{\sort}_d = \frac{1}{2(d+1)}l'_\varepsilon = 4\nu^2\left(1 + \frac{1}{\sin\theta^{\sort}_d}\right)\varepsilon^{1/\nu}$ and hence contains a ball of radius $\varepsilon^{1/\nu}$. As in the previous part, this fact means that there exists $x \in \mathcal{X}^{\sort}$ such that $B\left(x,(\varepsilon/\max\{1,\|f\|^2_{\mathcal{H}}\widetilde{C}_{\mathcal{K}, \nu, d}(1 + P_{\nu,d}\varepsilon^{1/\nu})\})^{1/\nu}\right) \subset B(x,\varepsilon^{1/\nu}) \subset \mathcal{X}^{\sort}$. This allows us to apply Lemma \ref{probabilityboundsforhlemma} for the tail bound of $h_{\sort X,\mathcal{X}^{\sort}}$, which completes the proof.
\end{proof}

\begin{figure}[ht!]
    \centering
    \includegraphics[width=0.45\linewidth]{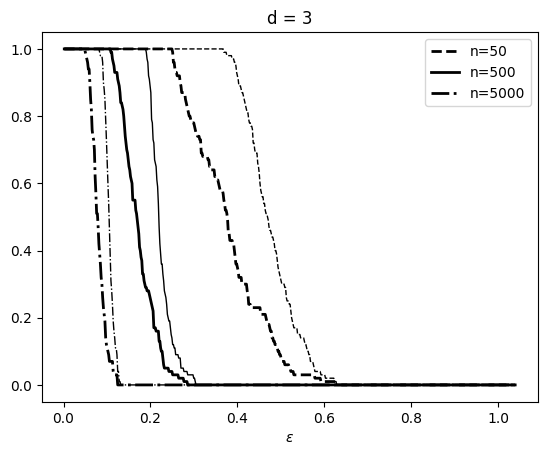}\hspace{1cm}
    \includegraphics[width=0.45\linewidth]{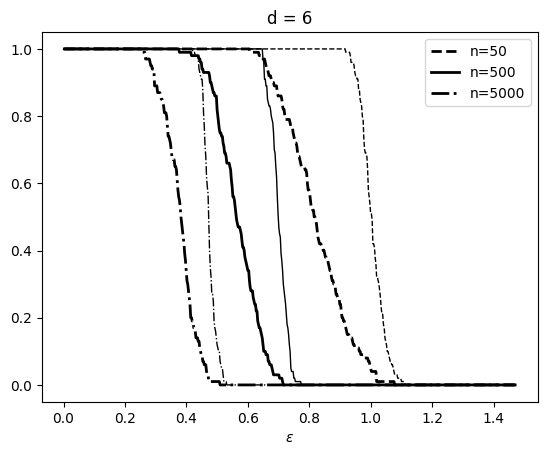}\\
    \includegraphics[width=0.45\linewidth]{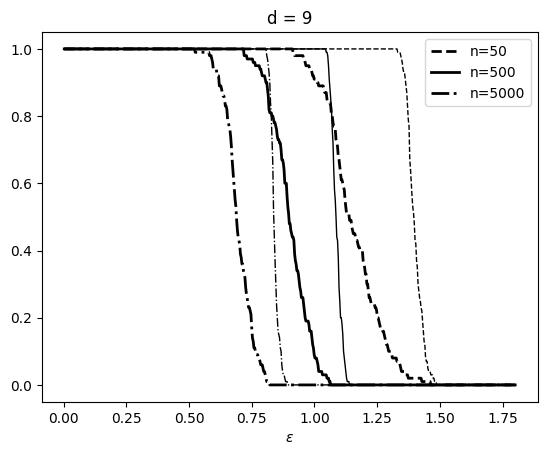}\hspace{1cm}
    \includegraphics[width=0.45\linewidth]{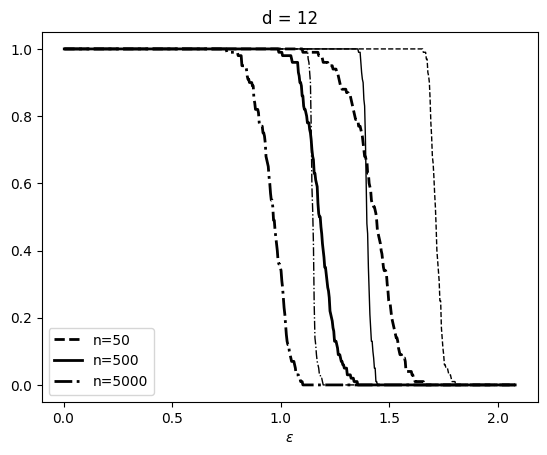}
    \caption{Numerical illustration of the probabilistic bounds $\mathbb{P}[h_{X, \mathcal{X}} > \varepsilon]$ (thin lines) and $\mathbb{P}[h_{\sort X, \mathcal{X}^{\sort}} > \varepsilon]$ (thick lines).}
    \label{fig:ph}
\end{figure}

To further illustrate the improvement from the sorting trick, we consider a numerical example of $h_{X, \mathcal{X}}$ and $h_{\sort X, \mathcal{X}^{\sort}}$ with the underlying $\mathbb{P}$ given by $Unif[0,1]^d$, under various $n$ and $d$. The results are summarized in Table \ref{Ehresults} and Figure \ref{fig:ph}.

\begin{table}[ht!]
\centering
\caption{Numerical illustration of the expected fill distances for $\mathcal{X}$ and $\mathcal{X}^{\sort}$.}
\label{Ehresults}
\begin{tabular}{ccccccc}
\toprule
\multirow{2}{*}{$d$} & \multicolumn{2}{c}{$n=50$} & \multicolumn{2}{c}{$n=500$} & \multicolumn{2}{c}{$n=5000$} \\
\cmidrule(lr){2-3} \cmidrule(lr){4-5} \cmidrule(lr){6-7}
 & {$\mathbb{E}\left[h_{X, \mathcal{X}}\right]$} & {$\mathbb{E}\left[h_{\sort X, \mathcal{X}^{\sort}}\right]$} & {$\mathbb{E}\left[h_{X, \mathcal{X}}\right]$} & {$\mathbb{E}\left[h_{\sort X, \mathcal{X}^{\sort}}\right]$} & {$\mathbb{E}\left[h_{X, \mathcal{X}}\right]$} & {$\mathbb{E}\left[h_{\sort X, \mathcal{X}^{\sort}}\right]$} \\
\midrule
3  & 0.4768 & 0.3783 & 0.2242 & 0.1721 & 0.1055 & 0.0801 \\
6  & 1.0046 & 0.8213 & 0.7001 & 0.5661 & 0.4737 & 0.3774 \\
9  & 1.3995 & 1.1549 & 1.0874 & 0.9005 & 0.8404 & 0.6947 \\
12 & 1.7143 & 1.4259 & 1.3985 & 1.1781 & 1.1482 & 0.9594 \\
\bottomrule
\end{tabular}
\end{table}


\section{Decay rates of eigenvalues}\label{edrsection}

Given a symmetric positive (semi)definite kernel $\mathcal{K} : \mathcal{X}\times\mathcal{X}\rightarrow \mathbb{R}$, we define the corresponding Hilbert-Schmidt operator $\mathcal{T} : L^2(\mathcal{X},\mathbb{P}) \rightarrow L^2(\mathcal{X},\mathbb{P})$ as follows:
\begin{equation*}
    \mathcal{T}(f) := \int_\mathcal{X}\mathcal{K}(.,x)f(x)\mathbb{P}(dx).
\end{equation*}
Mercer's theorem implies that $\mathcal{T}$ is positive (semi)definite and self-adjoint. We define similarly the operator $\mathcal{T}^{\sort}$ corresponding to the kernel $\mathcal{K}^{\sort}$. We denote the non-negative eigenvalues of $\mathcal{T}$ and $\mathcal{T}^{\sort}$ by
\begin{equation*}
    \lambda_1 \geq \lambda_2 \geq \cdots \geq \lambda_j \geq \cdots \geq 0, \quad \text{and} \quad \lambda^{\sort}_1 \geq \lambda^{\sort}_2 \geq \cdots \geq \lambda^{\sort}_j \geq \cdots \geq 0,
\end{equation*}
respectively.

\begin{theorem}\label{edrupperboundstheorem}
    Suppose that the probability measure $\mathbb{P}$ has an $S_d$-invariant smooth density bounded below by some $\underline{\rho} > 0$, and that the positive semidefinite kernel $\mathcal{K} \in C^{\nu}(\mathcal{X}\times \mathcal{X})$. Then
    \begin{align*}
        \lambda_j &\leq \frac{(\nu+d)!}{d!}\cdot\frac{C_{\mathcal{K},\nu}}{(2\pi)^\nu}\cdot\frac{((1+\nu)d^{\nu}\omega_d/\underline{\rho})^{\nu/d}}{j^{\nu/d}},\\
        \lambda^{\sort}_j &\leq \frac{(\nu+d)!}{d!}\cdot\frac{C_{\mathcal{K},\nu}}{(2\pi)^\nu}\cdot\frac{((1+\nu)d^\nu\omega_d/\underline{\rho})^{\nu/d}}{(d!)^{\nu/d-1}j^{\nu/d}},
    \end{align*}
where $\omega_d$ denotes the volume of a unit-radius $d$-ball.
\end{theorem}

The proof of Theorem \ref{edrupperboundstheorem} is given in the appendix.

\bigskip
The upper bounds in Theorem \ref{edrupperboundstheorem} are derived using the approach of \cite{volkov2024optimal}, which considers the Neumann problem and applies the Weyl's law. The bound $1/j^{\nu/d}$ is nearly optimal for the $C^\nu$-smoothness class of kernels, as shown in  \cite{volkov2024optimal}. 
Relative to the unsorted kernel, the sorted kernel is shown to reduce the upper bound by a factor of $(d!)^{\nu/d-1}$ when $\nu/d > 1$, i.e., when the kernel is sufficiently smooth for the given dimension $d$.

\printbibliography

\newpage
\appendix

\begin{center}
{\LARGE\bfseries SUPPLEMENTARY MATERIALS: Approximating invariant functions with the sorting trick is theoretically justified\par}
\vspace{0.5cm} 

{\large
Wee Chaimanowong\textsuperscript{1}, Ying Zhu\textsuperscript{2}\par} 
\vspace{0.5cm}

{\normalsize
\textsuperscript{1}The Chinese University of Hong Kong\par
\textsuperscript{2}University of California, San Diego\par}

\end{center}

\section{Supportive results and additional proofs}

\subsection*{Proof of Lemma \ref{coneconditionalternativelemma}}
\begin{proof}
    For any $x' \in \Omega_x$ and a unit vector $\xi\in \mathbb{R}^d$ such that $x' + \tau \xi \in \Omega_x$ for all $\tau \in [0,r]$, given any $p \in \pi_{\nu}(\Omega_x)$, Markov’s inequality for an algebraic polynomial gives us
    \begin{equation*}
        \left|\frac{d}{d\tau}p(x'+\tau \xi)\right| \leq \frac{2\nu^2}{r}\|p\|_{L^\infty(\Omega_{x})}
    \end{equation*} 
    for $\tau \in [0,r]$. Let $x^* \in \Omega_{x}$ be such that $p(x^*) = \|p\|_{L^\infty(\Omega_{x})}$, which is possible by the compactness of $\Omega_{x}$, where we canonically embed $\pi_{\nu}(\mathbb{R}^d)\ni p$ into $\pi_{\nu}(\Omega_{x})$ via restriction. By the $(\theta, r)$-interior cone condition, we can find $C(x^*, \xi_{x^*}, \theta, r) \subset \Omega_x$. Such a cone would contain a ball $B(y_{x^*},h)$, where $y_{x^*} := x^* + h\xi_{x^*}/\sin\theta$ (see \cite[Theorem 3.7]{wendland2004scattered}), and hence, there exists a data point $x_i \in B(y_{x^*},h) \subset C(x^*, \xi_{x^*}, \theta, r) \subset \Omega_x$. In particular, the index set $I_x$ is non-empty. It follows that
    \begin{multline*}
        |p(x^*) - p(x_i)| \leq  \int_0^{\|x^*-x_i\|_2}\left|\frac{d}{d\tau}p\left(x^* + \tau\frac{x^* - x_i}{\|x^* - x_i\|_2}\right)\right|d\tau \leq \frac{2\nu^2}{r}\|p\|_{L^\infty(\Omega_{t})}\cdot \|x^*-x_i\|_2\\
        \leq \frac{2\nu^2}{r}\|p\|_{L^\infty(\Omega_{x})}\cdot \left(\|x^*-y_{x^*}\|_2 + \|y_{x^*}-x_i\|_2\right) \leq \frac{2\nu^2}{r}\|p\|_{L^\infty(\Omega_{x})}\cdot \left(1 + \frac{1}{\sin\theta}\right) h \leq \frac{1}{2}\|p\|_{L^\infty(\Omega_{x})}.
    \end{multline*}
    In other words, we have $p(x_i) \geq \frac{1}{2}\|p\|_{L^\infty(\Omega_x)}$. Define a linear map $T : \pi_{\nu}(\Omega_{x}) \rightarrow L^\infty(I_x) \cong \mathbb{R}^{|I_{x}|}$ by $T(p) := [p(x_i)]_{i \in I_{x}}$. Then $T$ is injective and the inverse $T^{-1} : T(\pi_{\nu}(\Omega_{x})) \rightarrow \pi_{\nu}(\Omega_{x})$ has a bounded norm $\|T^{-1}\| = \sup_{p \in \pi_{\nu}(\Omega_{x})}\|p\|_{L^\infty(\Omega_{x})}/\|T(p)\|_{L^\infty(I_x)} \leq 2$. Consider a linear functional $\delta_{x} : \pi_{\nu}(\Omega_{x}) \rightarrow \mathbb{R}$, $\delta_xp := p(x)$ and $\widetilde{\delta}_x := \delta_x\circ T^{-1} : T(\pi_{\nu}(\Omega_{x}))\rightarrow \mathbb{R}$. Then $\widetilde{\delta}_x$ can be norm-preservingly extended by Hahn-Banach Theorem to the linear functional $\widetilde{\delta}_{x,ext} : \mathbb{R}^{|I_{x}|}\rightarrow \mathbb{R}$. Such a linear functional can be represented by an inner product, i.e. there exists $\{\widetilde{u}_i(x)\}_{i \in I_{x}}$ such that $\widetilde{\delta}_{x, ext}(v) = \sum_{i \in I_{x}}\widetilde{u}_i(x)v_i$ for any $v = (v_i)_{i\in I_{x}} \in \mathbb{R}^{|I_x|}$. For $i \in \{1,\cdots, n\} \setminus I_{x}$ we can take $\widetilde{u}_i(x) = 0$, which automatically implies that $\widetilde{u}_i(x) = 0$ if $\|x-x_i\|_2 > D$. To summarize, we have $\{\widetilde{u}_i(x)\}_{i=1}^n$ such that $\widetilde{u}_i(x) = 0$ provided that $\|x - x_i\|_2 > D$, 
    \begin{equation*}
        p(x) = \delta_x(p) = \widetilde{\delta}_{x,ext}\circ T(p) = \sum_{i\in I_{x}}\widetilde{u}_i(x)p(x_i) = \sum_{i=1}^n\widetilde{u}_i(x)p(x_i),
    \end{equation*}
    and $\sum_{i=1}^n|\widetilde{u}_i(x)| = \sum_{i\in I_{x}}|\widetilde{u}_i(x)| = \|\widetilde{\delta}_{x,ext}\|_{L^{\infty}(I_x)^*} = \|\widetilde{\delta}_{x}\|_{L^{\infty}(I_x)^*} \leq \|\delta_x\|\cdot \|T^{-1}\| \leq 2$, as stated. 
\end{proof}

\subsection*{Proof of Theorem \ref{thm:localapproxerror}:} 
\begin{proof}
Recalling $\hat{f}_n = \mathcal{K}(.,X)\mathcal{K}(X,X)^{-1}y$, we have
    \begin{multline*}\label{boundfatt}
        \left|f(x) - \hat{f}_n(x)\right| = \left|\left\langle f, \mathcal{K}(.,x) - \mathcal{K}(x,X)\mathcal{K}(X,X)^{-1}\mathcal{K}(.,X)\right\rangle_{\mathcal{H}}\right|\\
        \leq \left\| f \right\|_{\mathcal{H}} \cdot \left\|\mathcal{K}(.,x) - \mathcal{K}(x,X)\mathcal{K}(X,X)^{-1}\mathcal{K}(.,X)\right\|_{\mathcal{H}} = \|f\|_{\mathcal{H}}\cdot \sqrt{Q_x(u^*(x))},\numberthis
    \end{multline*}
where $Q_x : \mathbb{R}^n\rightarrow \mathbb{R}$ is given by
    \begin{equation*}
        Q_x(u) := \left\|\mathcal{K}(.,x) - \mathcal{K}(.,X)u\right\|_{\mathcal{H}}^2 = \mathcal{K}(x,x) - 2\mathcal{K}(x,X)u + u^{\intercal}\mathcal{K}(X,X)u
    \end{equation*}
and $u^*(x) = \mathcal{K}(X,X)^{-1}\mathcal{K}(X,x) \in \mathbb{R}^n$.
Using (\ref{boundfatt}), it remains for us to prove the following claim:

\bigskip

\emph{Given a positive definite kernel $\mathcal{K} \in C^{\nu}(\mathcal{X}\times \mathcal{X})$ and a set of sample points $\{x_i\}_{i=1}^n \subset \mathcal{X}$ and any $x \in \mathcal{X}$, $u^*(x) \in \mathbb{R}^n$ is a global minimum of $Q_x$ and $Q_x(u^*(x)) \leq 8{\nu + 2d \choose 2d}C_{\mathcal{K},\nu, x,D}D^{2\nu}$.}

\bigskip

The idea is to bound $Q_x(u^*(x))$ by $Q_x(\widetilde{u}(x))$ with $\widetilde{u}(x)$ given in Lemma \ref{coneconditionalternativelemma}, and use the reproducing properties of $\widetilde{u}(x)$ to simplify the Taylor expansion of $\mathcal{K}$ centered at $(x,x)$. Since $\mathcal{K}$ is positive definite and $Q_x$ is a convex function, any stationary points of $Q_x$ is the global minimum. It follows that $u^*(x)$ is the global minimum as we can check that
    \begin{equation*}
        \frac{\partial Q_t}{\partial u_k}(u^*(x)) = -2\mathcal{K}(x,x_k) + 2\mathcal{K}(x_k,X)u^*(x) = -2\mathcal{K}(x,x_k) + 2\mathcal{K}(x,X)\mathcal{K}(X,X)^{-1}\mathcal{K}(X,x_k) = 0.
    \end{equation*}
    Note that $\mathcal{K}(x,X)\mathcal{K}(X,X)^{-1}\mathcal{K}(X,x_k)$ is simply the orthogonal projection of $\mathcal{K}(.,x_k)$ to the space spanned by $\{\mathcal{K}(.,x_j)\}_{j=1}^n$, and thus is equal to $\mathcal{K}(.,x_k)$. Let $\left\{\widetilde{u}_j(x)\right\}_{j=1}^n$ be defined as in Lemma \ref{coneconditionalternativelemma}. Then
    \begin{multline*}
        Q_x(u^*) \leq Q_x(\widetilde{u}) = \mathcal{K}(x,x) - 2\sum_{j=1}^n\widetilde{u}_j(x)\mathcal{K}(x,x_j) + \sum_{i,j=1}^n\widetilde{u}_i(x)\widetilde{u}_j(x)\mathcal{K}(x_i,x_j)\\
        = \mathcal{K}(x,x) - 2\sum_{j=1}^n\widetilde{u}_j(x)\left(\sum_{|\alpha|< \nu}\frac{(x-s_j)^\alpha}{\alpha!}\partial^\alpha_2\mathcal{K}(x,x) + \sum_{|\alpha|=\nu}R_{0,\alpha}(x,x;x,x_j)(x-x_j)^\alpha\right)\\
        + \sum_{i,j=1}^n\widetilde{u}_i(x)\widetilde{u}_j(x)\Bigg(\sum_{|\alpha|+|\beta|<\nu}\frac{(x-x_i)^\alpha(x-x_j)^\beta}{\alpha!\beta!}\partial^\alpha_1\partial^\beta_2\mathcal{K}(x,x) + \sum_{|\alpha|+|\beta| = \nu}R_{\alpha, \beta}(x,x;x_i,x_j)(x-x_i)^\alpha(x-x_j)^\beta\Bigg)\\
        =  -2\sum_{j=1}^n\sum_{|\alpha|=\nu}\widetilde{u}_j(x)R_{0,\alpha}(x,x;x,x_j)(x-x_j)^\alpha\\
        + \sum_{i,j=1}^n\sum_{|\alpha|+|\beta|=\nu}\widetilde{u}_i(x)\widetilde{u}_j(x)R_{\alpha,\beta}(x,x;x_i,x_j)(x-x_i)^\alpha(x-x_j)^\beta.
    \end{multline*}
    The first inequality follows from the fact that $u^*$ is the global minimum. In the first equality, we use the fact that $\mathcal{K} \in C^{\nu}(\mathcal{X}\times \mathcal{X})$ and apply the multivariate Taylor's Theorem from any point $(x,x) \in \mathcal{X}\times \mathcal{X}$ to $(x,x_i)$ and $(x_i,x_j)$, and the last equality follows from the reproducing properties of $\widetilde{u}$ for $\pi_{2\nu}(\mathbb{R}^d)$ polynomials. We also denote the remainder by 
    \begin{equation*}
        R_{\alpha, \beta}(x_1,x_2;x'_1,x'_2) := \frac{1}{\alpha!\beta!}\partial^\alpha_1\partial^\beta_2\mathcal{K}(\xi_1, \xi_2)
    \end{equation*}
    for some $(\xi_1, \xi_2) \in \mathcal{X}\times \mathcal{X}$ on the line connecting any $(x_1, x_2)\in \mathcal{X}\times \mathcal{X}$ and $(x'_1, x'_2)\in \mathcal{X}\times \mathcal{X}$. From the vanishing property, $\widetilde{u}_j(x) = 0$ if $\|x-x_j\|_2 > D$, so we have $\left|x-x_i\right|^\alpha \leq \|x-x_i\|_2^{|\alpha|} \leq D^{|\alpha|}$. Hence, we can assume that $x_i \in B(x, D)$ for any $i$ that is relevant to us, and therefore,
    \begin{equation*}
        |R_{\alpha, \beta}(x,x;x_i,x_j)| \leq \max_{\xi_1, \xi_2 \in \mathcal{X}\cap B(x,D)}\max_{|\alpha|+|\beta|=\nu}\frac{1}{\alpha!\beta!}\left|\partial_1^\alpha\partial_2^\beta\mathcal{K}(\xi_1,\xi_2)\right| =: C_{\mathcal{K},\nu, x,D},
    \end{equation*}
    which is finite because $\mathcal{X}$ is compact. 
    Continuing the above calculation with the triangle inequality yields
    \begin{multline*}
        Q_x(u^*(x)) \leq 2\cdot {\nu + d \choose d} \cdot C_{\mathcal{K},\nu, x,D}D^{\nu}\sum_{j=1}^n\left|\widetilde{u}_j(x)\right| + {\nu + 2d \choose 2d}\cdot C_{\mathcal{K},\nu, x,D}D^{\nu}\left(\sum_{j=1}^n\left|\widetilde{u}_j(x)\right|\right)^2\\
        \leq 4\left({\nu + d \choose d}+{\nu + 2d \choose 2d}\right)C_{\mathcal{K},\nu, x,D}D^{\nu} \leq 8{\nu + 2d \choose 2d}C_{\mathcal{K},\nu, x,D}D^{\nu},
    \end{multline*}
    where we have used the property that $\sum_{j=1}^n\left|\widetilde{u}_j(x)\right| \leq 2$ to conclude the second last inequality, and ${\nu +d\choose d} \leq {\nu + 2d\choose 2d}$ in the final inequality to further simplify the expression.
\end{proof}

\subsection*{Proof of Lemma \ref{computeinteriorconelemma}}

\begin{proof}
    To verify the interior cone condition in Definition \ref{interiorconedefinition}, we only need to consider the most restrictive case where $x \in \mathcal{X}$ is one of the vertices of $\mathcal{X}$. For other cases, including when $x \in \mathcal{X}$ is an interior point, we have that $x$ is one of the vertices of $\frac{1}{2}\mathcal{X}$ embedded in $\mathcal{X}$ under some translation. Therefore, let us consider $x=(0,\cdots,0)$ and construct a cone in the direction $\xi_x$ from $x$ to the center of the cube $\bar{v} := \frac{1}{2}(1,\cdots,1)$. Projecting $\bar{v}$ orthogonally onto the $x^d = 0$ face of $\partial\mathcal{X}$ (or any other faces as we still have an equivalent result), we obtain $\proj \bar{v} = \frac{1}{2}(1,\cdots,1,0)$. The angle $\tilde{\theta}_d$ between $\bar{v}$ and $\proj \bar{v}$ can be computed as follows:
    \begin{equation*}
        \cos \tilde{\theta}_d = \frac{\bar{v}\cdot \proj\bar{v}}{\|\bar{v}\|_2\|\proj\bar{v}\|_2} = \sqrt{\frac{d-1}{d}} \ \implies \ \sin\tilde{\theta}_d = \frac{1}{\sqrt{d}}.
    \end{equation*}
    Therefore, to ensure the cone is contained in $\frac{1}{2}\Omega$, we choose $\theta_d = \tilde{\theta}_d = \arcsin 1/\sqrt{d}$, and $r_d = 1/2$.

\bigskip
    For $\mathcal{X}^{\sort}$, the vertices are $v_k = (\underbrace{1,\cdots,1}_k,0,\cdots,0)$. We again consider the most restrictive case where $x = v_0$, and let us construct a cone in the direction $\xi_x$ from $x$ to the simplex's center of gravity
    \begin{equation*}
        \bar{v} := \frac{1}{d+1}\sum_{k=0}^d v_k = \frac{1}{d+1}\left(d,d-1,\cdots,2,1\right).
    \end{equation*}
    Any faces of $\partial\Omega^{\sort}$ that contain $v_0$ are embedded in one of the subspaces $\mathcal{L}_k := span\{v_l\}_{l=1}^d\setminus\{v_k\}$. By considering the angle between $\bar{v}$ and its orthogonal projection
    \begin{equation*}
        \proj_k\bar{v} = \begin{cases}\frac{1}{d+1}\left(\frac{2d-1}{\sqrt{2}},\frac{2d-1}{\sqrt{2}},d-2,\cdots,2,1\right), & k=1\\
        \frac{1}{d+1}\left(d,d-1,\cdots,d-k+2,\frac{2d-2k+1}{\sqrt{2}},\frac{2d-2k+1}{\sqrt{2}},d-k-1,\cdots,2,1\right), & k=2,\cdots,d-1\\
        \frac{1}{d+1}(d,d-1,\cdots,2,0), & k=d
        \end{cases}
    \end{equation*}
    on each $\mathcal{L}_k$, we find that for all sufficiently large $d$, the angle $\tilde{\theta}^{\sort}_d$ which decays the fastest
    is between $\bar{v}$ and $\proj_d\bar{v}$:
    \begin{equation*}
        \cos \tilde{\theta}^{\sort}_d = \frac{\bar{v}\cdot \proj_d\bar{v}}{\|\bar{v}\|_2\|\proj_d\bar{v}\|_2} = \sqrt{\frac{d(d+1)(2d+1)-6}{d(d+1)(2d+1)}} \ \implies \ \sin\tilde{\theta}^{\sort}_d = \sqrt{\frac{6}{2d^3+3d^2+d}} \geq \frac{1}{d^{3/2}}.
    \end{equation*}
    In terms of $r^{\sort}_d$, we study the shortest distance between $\bar{v}$ and any faces of $\partial\mathcal{X}^{\sort}$ and obtain
    \begin{multline*}
        \|\bar{v}-\proj_d\bar{v}\|_2 = \frac{1}{d+1} < \min_{k=1,\cdots,d-1}\|\bar{v}-\proj_k\bar{v}\|_2\\
        = \min_{k=1,\cdots,d-1}\sqrt{\left(d-k+1-\frac{2d-2k+1}{\sqrt{2}}\right)^2+\left(d-k-\frac{2d-2k+1}{\sqrt{2}}\right)^2}.
    \end{multline*}
    Therefore, to ensure the cone is contained in $\frac{1}{2}\mathcal{X}^{\sort}$, we choose $\theta^{\sort}_d := \arcsin 1/d^{3/2}$, and $r^{\sort}_d := \frac{1}{2(d+1)}$.
\end{proof}

\subsection*{Proof of Theorem \ref{edrupperboundstheorem}}
\begin{proof}
    Our proof is based on \cite{volkov2024optimal} with an application of Weyl's law for a Riemannian manifold with boundary. In our case, we can define the metric tensor on $\mathcal{X}$ by $g(x) := \sum_{a=1}^d p(x)^{2/d}\left(dx^a\right)^2$, where $p$ denotes the probability density of $\mathbb{P}$. Note that $g$ is guaranteed to be positive-definite by the lower bound $p(x) > \underline{\rho}$, and we have $p(x) = \sqrt{|\det g(x)|}$. The theory of elliptic operators is well established when the domain boundary is smooth (i.e. $C^\infty$); however, we are interested in cases where $\partial\mathcal{X}$ is not necessarily smooth. To fix this issue, we consider a sequence $\{\mathcal{X}_k\}_{k=1}^\infty$ such that $\partial\mathcal{X}_k$s are $C^\infty$, $\mathcal{X}_1\subset \mathcal{X}_2\subset \cdots\subset \mathcal{X}$, and $\mathcal{X} = \bigcup_{k=1}^\infty\mathcal{X}_k$, as described in \cite{antonini2024smooth}, and define $\mathcal{T}_k : L^2(\mathcal{X},\mathbb{P})\rightarrow L^2(\mathcal{X},\mathbb{P})$ by $\mathcal{T}_kf := \int_{\mathcal{X}_k}\mathcal{K}(.,x)f(x)\mathbb{P}(dx)$. Then from the Dominated Convergence Theorem, we obtain
    \begin{equation*}
        \|\mathcal{T}-\mathcal{T}_k\| \leq \left[\int_{\mathcal{X}}\int_{\mathcal{X}}|\mathcal{K}(x_1,x_2)|^2(1-1_{x_2\in \mathcal{X}_k})\mathbb{P}(dx_1)\mathbb{P}(dx_2)\right]^{1/2}\rightarrow 0.
    \end{equation*}
    Thus, we have a sequence $\{\mathcal{T}_k\}_{k=1}^\infty$ of compact operators converging to $\mathcal{T}$, and by \cite[Chapter XI.9, Lemma 5]{dunfort1963linear}, $s_j(\mathcal{T}) = \lim_{k\rightarrow \infty}s_j(\mathcal{T}_{k})$. The remainder of the proof shows that $\{s_j(\mathcal{T}_{k})\}_{j=1}^\infty$ obeys the bound in the theorem, so the same must be true for $\{\lambda_j = s_j(\mathcal{T})\}_{j=1}^\infty$. Therefore, in what follows, let us proceed as if $\partial\mathcal{X}$ is smooth. 

    \bigskip

We consider the Laplace-Beltrami operator $\Delta : C^\infty(\mathcal{X})\rightarrow C^\infty(\mathcal{X})$. It is known that the solution $\{f_j \in C^\infty(\mathcal{X})\}_{j=0}^\infty$ to the Neumann problem 
    \begin{equation*}
        \Delta f + \mu_j f = 0, \quad \nabla f|_{\partial\mathcal{X}}\cdot n_\mathcal{X} = 0, \quad \|f\|_{L^2(\mathcal{X},\mathbb{P})} = 1
    \end{equation*}
    gives an orthonormal basis for $L^{2}(\mathcal{X},\mathbb{P})$ (see e.g. \cite{ito1961neumann}). The corresponding eigenvalues are $\{\mu_j\}_{j=0}^\infty$, satisfying $\mu_0=0 < \mu_1\leq \mu_2 \leq\cdots$. In particular, it is easy to see that $f_0 = 1$ is the $\mu_0 = 0$ eigenfunction. Asymptotically in $j$, Weyl's law \cite{ivrii1980second} gives that
    \begin{equation*}
        \mu_j \sim \frac{4\pi^2}{(\omega_d|\mathcal{X}|)^{2/d}}\cdot j^{2/d}.
    \end{equation*}
Before we proceed to the main proof, let us fix some notation. For $s\geq 0$, we denote the Sobolev Hilbert space by
\begin{equation*}
    H^s(\mathcal{X},\mathbb{P}) := \left\{f = \sum_{j=0}^\infty a_jf_j\ |\ \sum_{j=0}^\infty\max\{\mu_1,\mu_j\}^s|a_j|^2 < \infty\right\}
\end{equation*}
with inner product
\begin{equation*}
    \left\langle \sum_{j=0}^\infty a_jf_j, \sum_{j=0}^\infty b_jf_j\right\rangle_{H^s(\mathcal{X}, \mathbb{P})} := \sum_{j=0}^\infty \max\{\mu_1,\mu_j\}^sa_jb_j.
\end{equation*}
Note that we have $H^0(\mathcal{X},\mathbb{P}) = L^2(\mathcal{X},\mathbb{P})$. 

\bigskip
Given a compact operator $\mathcal{C}:\mathcal{H}_1\rightarrow \mathcal{H}_2$, we denote the singular values of $\mathcal{C}$ by $\{s_j(\mathcal{C})\}_{j=1}^\infty$, i.e., the eigenvalues of $\sqrt{\mathcal{C}^*\mathcal{C}}$. We adopt the convention that singular values are ordered in descending order: $s_1(\mathcal{C}) \geq s_2(\mathcal{C}) \geq \cdots$. Note that if $\mathcal{C}$ is self-adjoint, then $s_j(\mathcal{C})$ are the eigenvalues of $\mathcal{C}$.

\bigskip

We define $\mathcal{T}_{i_1,\cdots,i_\nu} : L^2(\mathcal{X},\mathbb{P})\rightarrow L^2(\mathcal{X},\mathbb{P})$ by $\mathcal{T}_{i_1,\cdots,i_\nu}f := \int_{\mathcal{X}}\partial_{i_1}\cdots\partial_{i_\nu}\mathcal{K}(.,x)f(x)\mathbb{P}(dx)$. Since $\mathcal{T}_{i_1,\cdots,i_{\nu}}$ is invariant under the permutation of the indices $i_1,\cdots,i_\nu$, we write $\mathcal{T}_\alpha := \mathcal{T}_{i_1,\cdots,i_\nu}$ where $\alpha = (\alpha^1,\cdots,\alpha^d) \in \mathbb{Z}^d_{\geq 0}$, $\alpha^a := |\{l | i_l = a\}|$, so that $\mathcal{T}_\alpha f := \int_{\mathcal{X}}\partial_2^\alpha\mathcal{K}(.,x)f(x)\mathbb{P}(dx)$. 

    \bigskip

    Going back to the proof, let us define an operator $\mathcal{J}:L^{2}(\mathcal{X},\mathbb{P})\rightarrow H^{2}(\mathcal{X},\mathbb{P})\hookrightarrow H^1(\mathcal{X},\mathbb{P})$ with $\mathcal{J}f_j = \frac{1}{\max\{\mu_1,\mu_j\}}f_j$, and then uniquely extend it to a compact operator on $L^2(\mathcal{X},\mathbb{P})$ with singular values $s_j(\mathcal{J}) = \frac{1}{\max\{\mu_1,\mu_{j-1}\}^{1/2}}$. Similarly, we can uniquely extend $\Delta$ based on its action on the basis set $\{f_j\}_{j=0}^\infty$ to a bounded linear operator $H^2(\mathcal{X},\mathbb{P})\rightarrow L^2(\mathcal{X},\mathbb{P})$. In particular, $\Delta\mathcal{J} : L^2(\mathcal{X},\mathbb{P})\rightarrow L^2(\mathcal{X},\mathbb{P})$ is a bounded operator given by $\Delta \mathcal{J}\left(\sum_{j=0}^\infty a_jf_j\right) := -\sum_{j=1}^\infty\frac{\mu_ja_j}{\max\{\mu_1,\mu_j\}}f_j$, and $\mathcal{T}(-\Delta\mathcal{J})$ is a compact operator with $\mathcal{T}(-\Delta \mathcal{J})f_0 = 0$ and coincides with $\mathcal{T}$ on $span\{f_0\}^{\perp}\subset L^2(\mathcal{X},\mathbb{P})$. From \cite[Proposition 2.2]{volkov2024optimal}, we have
    \begin{equation}\label{insertdeltaj}
        s_j(\mathcal{T}) \leq s_{j-1}(\mathcal{T}|_{span\{f_0\}^{\perp}}) = s_{j-1}(\mathcal{T}(-\Delta\mathcal{J})).
    \end{equation}
    For any eigenfunction $f_j$, we have
    \begin{multline*}\label{predeltajtosumtd}
        \mathcal{T}(-\Delta \mathcal{J})f_j = -\int_{\mathcal{X}}\mathcal{K}(.,x)\Delta \mathcal{J}f_j(x)\mathbb{P}(dx)\\
        = \int_{\mathcal{X}}\nabla \mathcal{K}(.,x)\cdot \nabla \mathcal{J}f_j(x)\mathbb{P}(dx) -\int_{\mathcal{X}}\nabla \cdot \left(\mathcal{K}(.,x)\nabla \mathcal{J}f_j(x)\right)\mathbb{P}(dx)\\
        = \int_{\mathcal{X}}\nabla \mathcal{K}(.,x)\cdot \nabla \mathcal{J}f_j(x)\mathbb{P}(dx) - \int_{\partial\mathcal{X}}\mathcal{K}(.,x)\nabla \mathcal{J}f_j(x)\cdot n_{\mathcal{X}} = \int_{\mathcal{X}}\nabla \mathcal{K}(.,x)\cdot \nabla \mathcal{J}f_j(x)\mathbb{P}(dx)\\
        = \sum_{i_1=1}^d\int_{\mathcal{X}}\partial_{i_1}\mathcal{K}(.,x)\partial^{i_1}\mathcal{J}f_j(x)\mathbb{P}(dx),\numberthis
    \end{multline*}
    where the boundary integral vanishes due to the Neumann boundary condition. We can uniquely extend the derivative operator $\partial^{i} := \frac{1}{p(x)^{2/d}}\partial_i$ based on its action on the basis set $\{f_j\}_{j=0}^\infty$ to a bounded linear operator $\partial^{i}: H^1(\mathcal{X},\mathbb{P})\rightarrow L^2(\mathcal{X},\mathbb{P})$ given by $\partial^i\left(\sum_{i=0}^\infty a_jf_j\right) := \sum_{j=0}^\infty a_j\partial^if_j$ with an operator norm $\|\partial^{i}\| \leq 1/\underline{\rho}^{1/d}$. Let $\mathcal{D}^i := \partial^i\mathcal{J} : L^2(\mathcal{X},\mathbb{P})\rightarrow L^2(\mathcal{X},\mathbb{P})$. Then we have $s_j(\mathcal{D}^i) \leq \|\partial^i\|s_j(\mathcal{J}) \leq \frac{1}{\underline{\rho}^{1/d}\max\{\mu_1,\mu_{j-1}\}^{1/2}}$.

    \bigskip
    Since (\ref{predeltajtosumtd}) is an equality of bounded operators on the basis set of $L^2(\mathcal{X},\mathbb{P})$, we have the following equality of bounded operators on $L^2(\mathcal{X},\mathbb{P})$ space:
    \begin{equation}\label{deltajtosumtd}
        \mathcal{T}(-\Delta\mathcal{J}) = \sum_{i_1=1}^d\mathcal{T}_{i_1}\mathcal{D}^{i_1}.
    \end{equation}
    In fact, (\ref{insertdeltaj}) and (\ref{deltajtosumtd}) are valid if $\mathcal{T}$ is replaced by any Hilbert-Schmidt integral operators, including $\mathcal{T}_\alpha$. In conjunction with \cite[Proposition 2.4 and Proposition 2.5]{volkov2024optimal}, by repeating the application of (\ref{insertdeltaj}) and (\ref{deltajtosumtd}) for the singular value of the sum and product of compact operators, as many times as $\mathcal{K}$ is differentiable, starting from $j = d^\nu(1+\nu) j' \geq j_{\nu} := \left(d^\nu + \frac{d^\nu-d}{d-1}\right)(j'-1) + \frac{d^\nu-1}{d-1}+1$ for any $j'\geq 2$, we have 
    \begin{multline*}
        \lambda_j = s_j(\mathcal{T})\\
        \leq s_{j_\nu}(\mathcal{T}) \leq s_{j_{\nu}-1}(\mathcal{T}(-\Delta\mathcal{J})) = s_{j_{\nu}-1}\left(\sum_{i_1=1}^d\mathcal{T}_{i_1}\mathcal{D}^{i_1}\right)\leq \sum_{i_1=1}^ds_{(j_{\nu}+d-2)/d-(j'-1)}(\mathcal{T}_{i_1})s_{j'}(\mathcal{D}^{i_1})\\
        \leq \sum_{i_1=1}^ds_{j_{\nu-1}}(\mathcal{T}_{i_1})s_{j'}(\mathcal{D}^{i_1}) \leq \sum_{i_1=1}^ds_{j_{\nu-1}-1}(\mathcal{T}_{i_1}(-\Delta\mathcal{J}))s_{j'}(\mathcal{D}^{i_1})\\
        = \sum_{i_1=1}^ds_{j_{\nu-1}-1}\left(\sum_{i_2=1}^d\mathcal{T}_{i_1,i_2}\mathcal{D}^{i_2}\right)s_{j'}(\mathcal{D}^{i_1})\leq \sum_{i_1,i_2=1}^ds_{(j_{\nu-1}+d-2)/d - (j'-1)}\left(\mathcal{T}_{i_1,i_2}\right)s_{j'}(\mathcal{D}^{i_1})s_{j'}(\mathcal{D}^{i_2})\\
        \leq \sum_{i_1,i_2=1}^ds_{j_{\nu-2}}\left(\mathcal{T}_{i_1,i_2}\right)s_{j'}(\mathcal{D}^{i_1})s_{j'}(\mathcal{D}^{i_2}) \leq \cdots \leq \sum_{i_1,\cdots,i_\nu=1}^ds_{j'}(\mathcal{T}_{i_1,\cdots,i_\nu})s_{j'}(\mathcal{D}^{i_1})\cdots s_{j'}(\mathcal{D}^{i_\nu})\\
        \leq \frac{1}{\underline{\rho}^{\nu/d}\mu_{j'-1}^{\nu/2}}\sum_{i_1,\cdots,i_\nu=1}^ds_{j'}(\mathcal{T}_{i_1,\cdots,i_\nu}) = \frac{1}{\underline{\rho}^{\nu/d}\mu_{j'}^{\nu/2}}\sum_{\alpha\in \mathbb{Z}^d_{\geq 0}; |\alpha| = \nu}\frac{\nu!}{\alpha!}s_{j'}(\mathcal{T}_\alpha)\\
        \leq \frac{1}{\underline{\rho}^{\nu/d}\mu_{j'-1}^{\nu/2}}\cdot \nu! {\nu+d\choose d} C_{\mathcal{K},\nu} \sim \frac{(\nu+d)!}{d!}C_{\mathcal{K},\nu}d^{\nu^2/d}(1+\nu)^{\nu/d}\frac{\left(\omega_d|\mathcal{X}|\right)^{\nu/d}}{(4\pi^2)^{\nu/2}\underline{\rho}^{\nu/d}}\cdot j^{-\nu/d},
    \end{multline*}
    where we have used the fact that $\frac{j_\nu+d-2}{d}-(j'-1) = j_{\nu-1}$ and defined
    \begin{equation*}
        C_{\mathcal{K}, \nu} := \max_{x_1, x_2 \in \mathcal{X}}\max_{|\alpha|+|\beta|=\nu}\frac{1}{\alpha!\beta!}\left|\partial_1^\alpha\partial_2^\beta\mathcal{K}(x_1,x_2)\right| \geq \frac{1}{\alpha'!}s_1(\mathcal{T}_{\alpha'}) \geq \frac{1}{\alpha'!}s_{j'}(\mathcal{T}_{\alpha'})
    \end{equation*}
    for all $\alpha'\in \mathbb{Z}^d_{\geq 0}$ such that $|\alpha'| = \nu$. 

\bigskip
    Next, we consider the eigenvalues of $\mathcal{T}^{\sort} : L^2(\mathcal{X},\mathbb{P})\rightarrow L^2(\mathcal{X},\mathbb{P})$ given by
    \begin{equation*}
        \mathcal{T}^{\sort}f = \int_{\mathcal{X}}\mathcal{K}^{\sort}(.,t)f(t)\mathbb{P}(dt) = \int_{\mathcal{X}}\mathcal{K}(\sort .,\sort t)f(t)\mathbb{P}(dt).
    \end{equation*}
    Note that any eigenfunctions $f \in L^2(\mathcal{X},\mathbb{P})$ such that $\mathcal{T}^{\sort}f = \lambda f$ for some $\lambda \in \mathbb{R}\setminus\{0\}$ must be permutation invariant, since
    \begin{multline*}
        f(\sort x) = \frac{1}{\lambda}(\mathcal{T}^{\sort}f)(\sort x) = \frac{1}{\lambda}\int_{\mathcal{X}}\mathcal{K}^{\sort}(\sort x,x')f(x')\mathbb{P}(dx')\\
        = \frac{1}{\lambda}\int_{\mathcal{X}}\mathcal{K}^{\sort}(x,x')f(x')\mathbb{P}(dx') = \frac{1}{\lambda}(\mathcal{T}^{\sort}f)(x) = f(x).
    \end{multline*}
    Therefore, the eigenvalues of $\mathcal{T}^{\sort}$ coincides with those of $\widetilde{\mathcal{T}}^{\sort} : L^2(\mathcal{X}^{\sort},\mathbb{P})\rightarrow L^2(\mathcal{X}^{\sort},\mathbb{P})$ given by
    \begin{equation*}
        \widetilde{\mathcal{T}}^{\sort}f := d!\int_{\mathcal{X}^{\sort}}\mathcal{K}(.,x)f(x)\mathbb{P}(dx),
    \end{equation*}
    since $\widetilde{\mathcal{T}}^{\sort}f = \mathcal{T}^{\sort}f$ for all permutation invariant $f$. We obtain the bound for $s_j(\widetilde{\mathcal{T}}^{\sort}/d!)$ following the general analysis with $\mathcal{X}^{\sort}$ replacing $\mathcal{X}$. Since $\lambda^{\sort}_j = s_j(\widetilde{\mathcal{T}}^{\sort}) = d!\cdot s_j(\widetilde{\mathcal{T}}^{\sort}/d!)$, we obtain that
    \begin{multline*}
        \lambda^{\sort}_j \lesssim d!\cdot C_{\mathcal{K},\nu}d^{\nu+\nu^2/d}(1+\nu)^{\nu/d}\frac{\left(\omega_d|\mathcal{X}^{\sort}|\right)^{\nu/d}}{(4\pi^2)^{\nu/2}}\cdot j^{-\nu/d}\\
        = d!\cdot \frac{(\nu+d)!}{d!}C_{\mathcal{K},\nu}d^{\nu^2/d}(1+\nu)^{\nu/d}\frac{\left(\omega_d|\mathcal{X}|\right)^{\nu/d}}{(4\pi^2)^{\nu/2}\underline{\rho}^{\nu/d}}\cdot (d!j)^{-\nu/d}.
    \end{multline*}
    
\end{proof}

\end{document}